\documentclass[acmtog,nonacm]{acmart}

\AtBeginDocument{%
  \providecommand\BibTeX{{%
    \normalfont B\kern-0.5em{\scshape i\kern-0.25em b}\kern-0.8em\TeX}}}

\usepackage{graphicx, amsmath, caption, subcaption, multirow, overpic, textpos, pifont, adjustbox}

\usepackage{subcaption}
\expandafter\def\csname ver@subfig.sty\endcsname{}

\usepackage{makecell}
\usepackage{float}
\usepackage{stfloats}

\usepackage{amsfonts}
\usepackage{algorithm}
\usepackage{array}

\usepackage{tabularx}

\usepackage{textcomp}
\usepackage{stfloats}
\usepackage{url}
\usepackage{verbatim}
\usepackage{hyperref}
\usepackage{amsmath}
\usepackage{bm}
\usepackage{subfig}
\usepackage{enumitem,kantlipsum}
\UseRawInputEncoding
\usepackage{multirow}
\usepackage{colortbl}

\usepackage{algorithmic}

\usepackage{appendix}

\definecolor{lightBlue}{rgb}{.62109375,.75390625,.82421875}
\definecolor{lightPink}{rgb}{.914,.804,.816}

\definecolor{greendatasets}{rgb}{.914,.804,.816}

\usepackage{pifont}

\definecolor{darkgreen}{rgb}{0.0, 0.5, 0.0}
\definecolor{darkred}{rgb}{0.5, 0.0, 0.0}
\definecolor{citecolor}{HTML}{0071BC}
\definecolor{linkcolor}{HTML}{000000}
\definecolor{acceptcolor}{HTML}{74C219}
\definecolor{rejectcolor}{HTML}{DE1616}
\definecolor{qcolor}{HTML}{536872}
\definecolor{demphcolor}{RGB}{100,100,100}

\begin{document}

\title{Unified Local and Global Attention Interaction Modeling\\
for Vision Transformers}


\author{Tan Nguyen}
\affiliation{%
\institution{Columbia Universiy}
\country{USA}}
\author{Coy D. Heldermon}
\affiliation{%
\institution{University of Florida}
\country{USA}}
\author{Corey Toler-Franklin}
\affiliation{%
\institution{Columbia University, Barnard College}\country{USA}}

\renewcommand{\shortauthors}{Nguyen, T.  et al.}

\begin{abstract} We present a novel method that extends the self-attention mechanism of a vision transformer (ViT) for more accurate object detection across diverse datasets. ViTs show strong capability for image understanding tasks such as object detection, segmentation, and classification. This is due in part to their ability to leverage global information from interactions among visual tokens. However, the self-attention mechanism in ViTs are limited because they do not allow visual tokens to exchange local or global information with neighboring features before computing global attention. This is problematic because tokens are treated in isolation when attending (matching) to  other tokens, and valuable spatial relationships are overlooked. This isolation is further compounded by dot-product similarity operations that make tokens from different semantic classes appear visually similar. To address these limitations, we introduce two modifications to the traditional self-attention framework; a novel aggressive convolution pooling strategy for local feature mixing, and a new conceptual attention transformation to facilitate interaction and feature exchange between semantic concepts. Experimental results demonstrate that local and global information exchange among visual features before  self-attention significantly improves performance  on challenging object detection tasks and generalizes across multiple benchmark datasets and challenging medical datasets. We publish source code and a novel dataset of cancerous tumors (chimeric cell clusters).
\end{abstract}



\maketitle

\section{Introduction}\label{sec:introduction}
Recent object detection models~\cite{DBLP:journals/corr/RenHG015,DBLP:journals/corr/RedmonDGF15,DBLP:journals/corr/abs-1708-02002,DBLP:journals/corr/LiuAESR15,DBLP:journals/corr/abs-2005-12872,DBLP:journals/corr/abs-1808-01244,Lyu2022RTMDetAE,zhang2023dino,DBLP:journals/corr/abs-2201-02605} are able to capture robust, representative, high-level semantic features across diverse datasets for accurate localization and classification of objects. These architectures incorporate learning-based visual feature encoders that are critical for \emph{perception object detection}, the process of identifying and interpreting visual information to recognize objects. Transformer architectures are at the forefront of these models, achieving state-of-the-art results on many object detection benchmarks ~\cite{10.1007/978-3-031-20077-9_17,DBLP:journals/corr/abs-2005-12872,9710580,DBLP:journals/corr/abs-2111-09883,DBLP:journals/corr/abs-2102-12122,DBLP:journals/corr/abs-2201-00520,xia2023datspatiallydynamicvision}. One of the reasons transformer encoders have been successful at object detection lies in their ability to model long-range dependencies between visual elements through the attention mechanism. This capability makes them  well-suited for visual detection tasks, where understanding spatial relationships at different scales and ranges is essential. 

\vspace{10pt}
\noindent Despite its recent successes and broad adoption, the transformer self-attention mechanism has inherent limitations when operating on complex datasets where different semantic objects exhibit visually similar appearances. Medical datasets with cancerous tumors in tissue scans or tumors in brain MRI images are examples. Queries, keys, and values for objects from different classes can become indistinguishable. Consequently, the attention map struggles to focus on relevant regions, and thus spans indiscriminate attention to non-relevant objects. In the case of cancer tumor detection, failing to differentiate between visually similar but conceptually different tissues could lead to false positives, inaccurate diagnoses and unnecessary invasive procedures.

\vspace{10pt}
\noindent To address these limitations, we propose a method that extends the self-attention mechanism in Vision Transformers to enable feature tokens to interact at both local and global scales before self-attention is applied. Inspired by complementary properties in localized convolutional interactions ~\cite{DBLP:journals/corr/abs-2103-15808, DBLP:journals/corr/abs-2107-06263} and global attention, our technique facilitates feature exchange, and allows tokens to develop more complex and distinct representations based on their true semantic class.  The approach incorporates two neural modules before the global attention step: (1) \emph{Aggressive Convolutional Pooling} that iteratively applies depth-wise convolution and pooling operations to  allow each feature token to capture both local and global interaction, and (2) a \emph{Conceptual Attention Transformation} implemented by a novel Conceptual Attention Transformer that leverages high-level conceptual knowledge ~\cite{Wu2021VisualTransformer} through a novel backward flow attention mechanism to provide a global perspective that complements local convolution interactions.

\vspace{10pt}
\noindent We leverage convolution applied in early stages to produce more distinct, well-differentiated features that effectively reduce smoothing caused by isolated feature interactions at later stages within the self-attention mechanism. The enriched features are further refined using conceptual attention transformation with a unique projection layer that integrates the input with the semantic conceptual tokens~\cite{Wu2021VisualTransformer}. The results produce visual tokens with improved contextual understanding and feature representation. Ultimately, self-attention is applied to these distinctive features that are stronger aligned with their true semantic classes.  

\vspace{10pt}
\noindent Our Enhanced Interaction Vision Transformer architecture shows substantial performance improvement for object detection over state-of-the-art transformer models for a broad range of self-attention module formations. Our contributions include:

\begin{itemize}[noitemsep,topsep=0pt]
    \item A novel aggressive depth-wise convolutional pooling module  that combines local interactions with global interactions before self-attention (Section \ref{subsec:aggressive-attention-pooling}).  
    \item A new conceptual attention transformation with a unique projection layer that integrates model inputs with  semantic conceptual tokens for enhanced feature representation and interaction (Section \ref{subsec:conceptual-attention-transformation}) .
    \item Comparative analysis on benchmark object detection datasets with a focus on medical imaging datasets.
    \item A challenging new benchmark object detection dataset of cancerous tumors (chimeric cell clusters) with ground truth annotations.
\end{itemize}

\section{Background}
This section provides background information on aspects of transformer models and self attention that motivate our work.

\vspace{10pt}
\noindent Transformer-based object detectors have the flexibility to learn universal features without implicit constraints of inductive bias inherent to CNN-based models like translational equivariance and locality~\cite{DBLP:journals/corr/abs-2010-11929}. Self attention, the fundamental mechanism of operation for  transformers~\cite{DBLP:journals/corr/VaswaniSPUJGKP17, DBLP:journals/corr/abs-2010-11929} effectively captures global information, granting each feature token a global receptive field. This capability is crucial for object detection, enabling the model components responsible for high level vision tasks to comprehend spatial relationships and extract meaningful spatial semantics for accurate detection. Several studies, which we discuss in Section~\ref{sec:related-work}, are relevant to our investigation of the effectiveness of transformer-based encoders for object detectors.

\vspace{10pt}
\noindent However, there are challenges. In multi-headed attention, feature tokens are projected through a linear aggregation along the channel-wise dimension to compute queries, keys, and values for self-attention (Equation \ref{eq:qkv-computation}). This can lead the network to rely more on positionally encoded information rather than extracting robust and representative features for downstream tasks where objects appear visually similar. In this formulation, feature tokens are treated in isolation when attending to others. Thus, for visually similar objects, their corresponding queries, keys, and values become nearly identical. 

\begin{equation} \label{eq:qkv-computation}
\begin{aligned}
K &= W_K \cdot X \\
V &= W_V \cdot X \\
Q &= W_Q \cdot X \\
attn= \sigma& \left( \frac{Q \cdot K^T}{\sqrt{d_k}} \right) \cdot V
\end{aligned}
\end{equation}

\noindent Where the matrices $W_K$, $W_V$, and $W_Q$ represent linear projection matrices without bias terms. The function $\sigma$ is applied to the result of the matrix multiplication between the query matrix $Q$ and the key matrix $K$, scaled by $\sqrt{d_k}$. This scaling factor normalizes the result based on the dot product in the channel dimension of size $d_k$ for both $Q$ and $K$, resulting in the attention matrix $attn$. The attention matrix $attn$ quantifies the relevance of each feature in the value vector $V$ for every query in $Q$.

 \vspace{10pt}
\noindent Vision transformers, the focus of our work, are designed for high-level computer vision operations on image data. ViTs scale efficiently when trained on large volumes of data. Thus, pre-trained ViTs are good foundational models able to transfer information learned from extensive datasets for detection tasks on mid-size and small image recognition benchmark datasets with prediction rates comparable to state-of-the-art CNN models~\cite{DBLP:journals/corr/abs-2010-11929,10378323,anonymous2024sam,Ranzinger10.1109}. We aim to use these models as a starting point for a technique that enhances feature representation for robust object detection in complex datasets where the target object is ambiguous (concealed-object datasets).

\section{Previous Work}\label{sec:related-work}
\noindent In this section, we review prior work, with a focus on object detection methods closely related to our approach. 

\vspace{10pt}
\noindent \emph{Object detection networks} are broadly categorized into multi-stage detectors~\cite{DBLP:journals/corr/GirshickDDM13,DBLP:journals/corr/Girshick15, DBLP:journals/corr/RenHG015,DBLP:journals/corr/HeZR014,8578742} and one-stage detectors ~\cite{DBLP:journals/corr/RedmonDGF15,DBLP:journals/corr/LiuAESR15,DBLP:journals/corr/abs-1708-02002}. Both approaches rely on a feature extraction to capture high-level semantic features that represent a variety of objects. Before the advent of transformers, the original approaches developed efficient convolution-based feature extractors as visual encoders tailored for object detection tasks~\cite{DBLP:journals/corr/HeZRS15, DBLP:journals/corr/abs-2003-13678, Pham2023NVAutoNetFA, Simonyan2014VeryDC, Tan2019EfficientNetRM}. Today, these designs complement and enhance overall performance in transformer-based architectures~\cite{DBLP:journals/corr/abs-2103-15808}. In multi-stage detectors, features are processed by an additional region proposal network (RPN) ~\cite{DBLP:journals/corr/RenHG015} that generates a set of potential regions of interest.   Features corresponding to these regions are then pooled~\cite{DBLP:journals/corr/LinDGHHB16} to incorporate multi-scale feature representations before performing the final detection. One-stage detectors bypass the RPN and directly generate detection anchors to simultaneously classify and localize objects. Recent autoregressive decoder methods~\cite{DBLP:journals/corr/abs-2005-12872} further eliminate the reliance on RPNs and anchor generation. These transformer-based decoders bypass the need for traditional non-maximum suppression, previously essential for both multi-stage and one-stage detectors.

\vspace{10pt}
\noindent \emph{Transformer-based Object Detection:} Previous studies on vision transformers predominantly utilize the vanilla self-attention mechanism, defined using queries, keys, and values (Equation \ref{eq:qkv-computation}). This mechanism often generates token-wise attention maps that exhibit excessive uniformity ~\cite{DBLP:journals/corr/abs-2106-03714}. This uniformity leads to dense aggregation of patch embeddings, that results in overly similar token representations—a phenomenon we refer to as the smoothing effect in self-attention. This effect is particularly pronounced in medical datasets, where objects from different classes often appear visually similar, and in natural datasets involving concealed or camouflaged objects. Prior works ~\cite{DBLP:journals/corr/abs-2106-03714, 10.1609/aaai.v37i11.26608} addressed this issue by enriching the attention maps after the self-attention computation. In this work, we assert that adding additional context to  feature representations prior to self-attention is an essential complementary step. This strategy enhances the expressiveness of attention maps, optimizes the aggregation of value features, and significantly improves overall performance.

\begin{figure*}[ht]
    \centering
    \includegraphics[width=\textwidth]{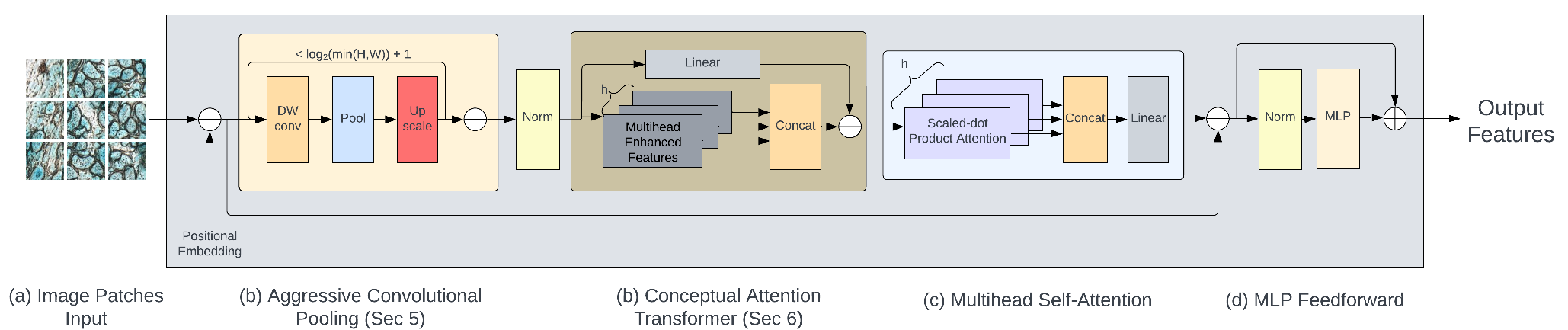}

    \caption{Overview of the Enhanced Interaction Vision Transformer: (a) The input image is tokenized using patch embedding to initialize the network. (b) The Aggressive Convolutional Pooling module (ACP) (Sec.~\ref{subsec:aggressive-attention-pooling}) iteratively enriches features by expanding the receptive field through convolution and pooling operations. The resulting features are normalized via layer normalization and passed to the Conceptual Attention Transformer (CAT) (Sec.~\ref{subsec:conceptual-attention-transformation}), which refines their representation through global interactions with high-level concept classes. (c) The refined features are processed through a multi-head self-attention mechanism, followed by (d) MLP feedforward layers, to generate the final output features of the enhanced interaction transformer block.
 }
    \label{fig:enhanced-vision-transformer-encoder}
\end{figure*}

\vspace{10pt}
\noindent \emph {Vision Transformers:} Recent advancements that adopt transformers for vision tasks have significantly enhanced the effectiveness of Vision Transformers as object detectors. Standard ViT object detection models~\cite{10.1007/978-3-031-20077-9_17} use a straight forward adaptation of a transformer model with minimal modifications. These original models achieve competitive results. Subsequent research improved feature extraction capabilities. The Swin Transformer~\cite{9710580, DBLP:journals/corr/abs-2111-09883} introduced a shifted window mechanism to reduce the computational overhead of global attention and incorporated a hierarchical structure for learning multi-scale object features. Inspired by the integration of convolutional properties to complement global attention~\cite{DBLP:journals/corr/abs-2103-15808},  DAT~\cite{DBLP:journals/corr/abs-2201-00520}, DAT++~\cite{xia2023datspatiallydynamicvision}, UniNet~\cite{Liu2021UniNetUA}, and EdgeNeXt~\cite{10.1007/978-3-031-25082-8_1} leverage both convolutional and transformer blocks.  We explore  related work  that optimizes the foundational structure of the ViT architecture for complex visual perception tasks.

\vspace{10pt}
\noindent \emph {Enhanced Feature Representations:} Our techniques are inspired by object detection methods that aim to improve feature representation in vision transformers~\cite{DBLP:journals/corr/abs-2107-06263}. These methods focus on feature representation after self attention outside of the encoder decoder modules. We argue that refinement after self-attention is prone to over-smoothing and is less effective. We generate more complexities before self attention, and use CNNs to add additional local perception that compliments global attention.

Other techniques reformulate self-attention to address limitations of transformers like  high computational cost and scalability. Deformable Attention mechanisms ~\cite{DBLP:journals/corr/abs-2201-00520, xia2023datspatiallydynamicvision, DBLP:journals/corr/abs-2010-04159} facilitate attention on relevant features and Neighborhood Attention ~\cite{10205440} introduces inductive biases like locality and translational equivariance. Refiner~\cite{DBLP:journals/corr/abs-2106-03714} and high-level concept attention ~\cite{Wu2021VisualTransformer} are enhancements that optimize attention mechanisms for feature diversity that improves performance across a broader range of detection tasks.
\vspace{4pt}

\vspace{10pt}
\noindent \emph{Concealed Object Detection:} Our work aligns with concealed object detection (COD) techniques for complex datasets. COD identifies objects that blend seamlessly  with the background, making them difficult to distinguish. A CNN-based approach~\cite{zhang2020multiscale} designed for cancer tumor detection in complex medical datasets incorporated local and background context while modifying the effective receptive field at different layers in a CNN to detect objects with non-discriminative features over a broad range of varying scales in a single forward pass.  SINet~\cite{DBLP:journals/corr/abs-2102-10274} introduced a search submodule combined with a texture enhanced module (TEM) to improve discriminative feature representations. These features are then used to generate a coarse attention map, enabling precise COD through a cascaded framework. SurANet~\cite{kang2024suranetsurroundingawarenetworkconcealed} incorporated surrounding environmental context during feature extraction, and applied a contrastive loss term, highlighting the benefits of increasing the complexity of visual features by fusing surrounding information. More recently, SAM-Adapter~\cite{Chen2023SAMFT} and SAM2-Adapter~\cite{Chen2024SAM2AdapterE} leveraged features from an  image encoder and mask decoder while integrating task-specific information through multi-layer perceptron adapters for successful COD.  Transformer-based architectures have consistently outperformed state-of-the-art methods for concealed object detection by exploiting the ability to model global relationships.

\section{Overview}

\noindent Figure \ref{fig:enhanced-vision-transformer-encoder} illustrates our system architecture adapted for the baseline (standard) vision transformer module. Two distinct interaction modules, \emph{Aggressive Convolutional Pooling} (ACP)  and \emph{Conceptual Attention Transformation} (CAT) enable feature tokens to interact before self-attention. These modules may be integrated into a wide range of ViT architectures. We position aggressive convolutional pooling before the conceptual attention transformation unit as convolutional operations utilize local kernels that capture localized interactions that complement the global attention mechanism. Convolutional properties enhance feature complexity early in the process, reducing smoothing effects when global attention is applied.   This additional enhancement transforms the input feature maps so that queries, keys, and values represent distinct features that encode their interrelationships. Enhanced feature complexity during the dot-product similarity within the attention mechanism leads to more easily detected differences between visually similar objects across different semantic classes.

\begin{figure*}[ht]
    \centering
    \includegraphics[width=\textwidth]{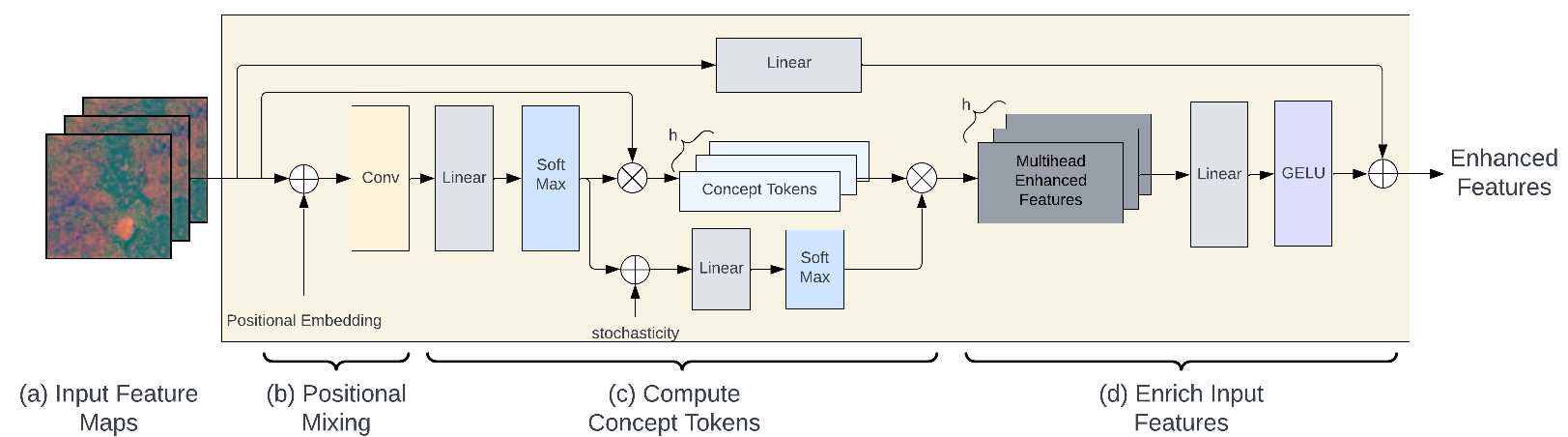}
    \caption{Conceptual Attention Transformation (CAT) Module: (a) CAT operates on input feature maps \(f \in \mathbb{R}^{C \times H \times W}\). (b) Positional mixing is first applied to integrate positional attributes into the input features. (c) These features are then used to compute global concept tokens. The computed global concepts are subsequently used to enhance the input feature (d), producing enhanced feature maps with global interaction information of the same shape as the input \(f\).
 }
    \label{fig:conceptual-attention-transformation}
\end{figure*}
\section{Aggressive Attention Pooling}\label{subsec:aggressive-attention-pooling}
\noindent We propose a novel aggressive depth-wise convolutional pooling layer before self-attention to enhance feature representations with both local and global context. Our strategy begins with the depth-wise convolution operation (DWConv) in Equation \ref{eq:lpu} from Local Perception Units (LPUs)~\cite{DBLP:journals/corr/abs-2107-06263,DBLP:journals/corr/abs-2201-00520,xia2023datspatiallydynamicvision}, and extends it with an iterative  pooling scheme that significantly increases the effective respective field for global interactions, rather than local-only operations that occur within the convolutional kernel's window. Like LPUs, our aggressive attention pooling method occurs before multi-head self-attention (MHSA).

\begin{equation}\label{eq:lpu}
    LPU(X) = \text{DWConv}(X) + X
\end{equation}
\begin{equation}\label{eq:lpu2}
    X_{out} = \text{MHSA}(\omega)
\end{equation}

 \noindent where $\omega$ is the output from the $LPU$ operation.

\vspace{10pt}
\noindent Let $X_0 \in \mathbb{R}^{C_0 \times H_0 \times W_0}$ denote the initial input feature map. At each step, $X_{i+1} \in \mathbb{R}^{2C_i \times H_i/2 \times W_i/2}$, where the spatial dimensions are reduced by a factor of $2$ and the channel size is doubled to preserve information.  Using Equation~\ref{eq:lpu}, we iteratively apply a sequence of convolutions with small kernel sizes interspersed with our additional max pooling operations. Our modification generalizes the LPU by repeating the LPU operation $n_{lpu} \leq \log_2(\min(H, W))$ times, where $H$ and $W$ represent the height and width of the feature map, respectively. Iterations beyond this threshold reduce one of the spatial dimensions to one. Average pooling is then applied to the spatial dimension to compute the final feature map $X_f \in \mathbb{R}^{C_f \times 1 \times 1}$, where $C_f \leq C_0 \cdot 2^{\log_2(\min(H_0, W_0))}$. The result is a feature map with a global receptive field. Conversely LPUs have smaller receptive fields limited to $k \times k$, where $k$ is the convolution kernel size.  Algorithm~\ref{alg:one} summarizes the process, where $X$ is the model input, $upscale$ is a small super-resolution network, and $out$ is a linear projection for $X$, and stores the summation of all feature maps at different scales.

\begin{algorithm} 
\caption{Aggressive Convolutional Pooling}
\begin{algorithmic}[1]
    \FOR{$i = 1$ to $n_{lpu}$}
        \STATE $X \gets \text{LPU}(X)$
        \STATE $\text{out} \gets \text{out} + \text{upscale}(X)$
        \STATE $X \gets \text{downscale}(X)$
    \ENDFOR
    \STATE $x \gets \text{MHSA}(\text{out})$

\end{algorithmic}
        \label{alg:one}
\end{algorithm}

\vspace{10pt}
  \noindent Let us denote $f_{cnn}^{i}$ as the convolutional operation at step $i$ that computes a new feature map $x^{i}$ of shape $(C \cdot 2^i, H / 2^i, W / 2^i)$ from an input feature of shape $(C, H, W)$. The function $f_{cnn}^{i}$ comprises a sequence of operations in the following order: Convolution with kernel size $3 \times 3$, ReLU activation function, and Max pooling with kernel size $2 \times 2$.

\vspace{10pt}
\noindent With this aggressive pooling layer, we obtain $M \leq \log_2(\min(H, W))$ feature maps. These $M$ features provide a global understanding of the surrounding context. It is important to note that each subsequent feature map $x^{i+1} = f_{cnn}^{i}(x^{i})$ has twice the receptive field of $x^{i}$. Consequently, the final feature map $x^{M}$ possesses the largest receptive field which can approximate the global receptive field when $M = \log_2(\min(H, W))$. We integrate these features with the input features, to gain combined local and global information at different scales in a framework inspired by the Feature Pyramid Network (FPN) architecture ~\cite{DBLP:journals/corr/LinDGHHB16}.

\vspace{10pt}
\noindent Finally, we aggregate all $x^{i}$, where $i \in \{1, ..., M\}$, into the input $x^{0}$. A naive approach would first perform concatenation, and then use a linear projection to map the concatenated feature into a $C$-dimensional feature space. However, we found this approach to be suboptimal, causing a large memory usage footprint. Instead, we propose  a mathematically equivalent memory-efficient alternative that computes the sum of individual linear projections for each $x^{i}$. The linear projection is used to map the $C \cdot 2^i$ channels of features into $C$ channels. To resolve mismatched spatial shapes, we employ a sequence of upscaling modules, each increasing the spatial resolution by a factor of $2$, followed by a convolution operation. Thus, each block consists of an upscaling step followed by a convolution. At most, $\log_2(\min(H, W))$ such blocks are required to restore the pooled features to their original size. We apply nearest tensor neighbor interpolation to address potential shape mismatches caused by max pooling when the spatial resolution is not divisible by $2$ as expressed as in Equation \ref{eq:aggressive-convolutional-pooling}:

\begin{equation} \label{eq:aggressive-convolutional-pooling}
    Y = f^0_{proj}(x^0) + \sum_{i=1}^{M} f_{upscale}^i\left(f_{cnn}^{i}(x^{i-1}) \right) 
\end{equation}

\noindent where $f_{upscale}$ is a small upscaling network that performs a sequence of convolution and upscaling operations to restore both the original channels and spatial dimensions for smaller feature maps.

\vspace{10pt}
\noindent ACP enables efficient global interaction, allowing tokens to acquire broader contextual information. Although each convolutional step operates within a local context, the successive extraction of local features culminates in a global receptive field. Integration of local and global information provides a robust foundation for subsequent self-attention computations, enhancing the model's ability to differentiate between semantically distinct tokens, even when they appear visually similar.

\begin{figure}[t]
    \centering
    \includegraphics[width=0.5
    \textwidth]{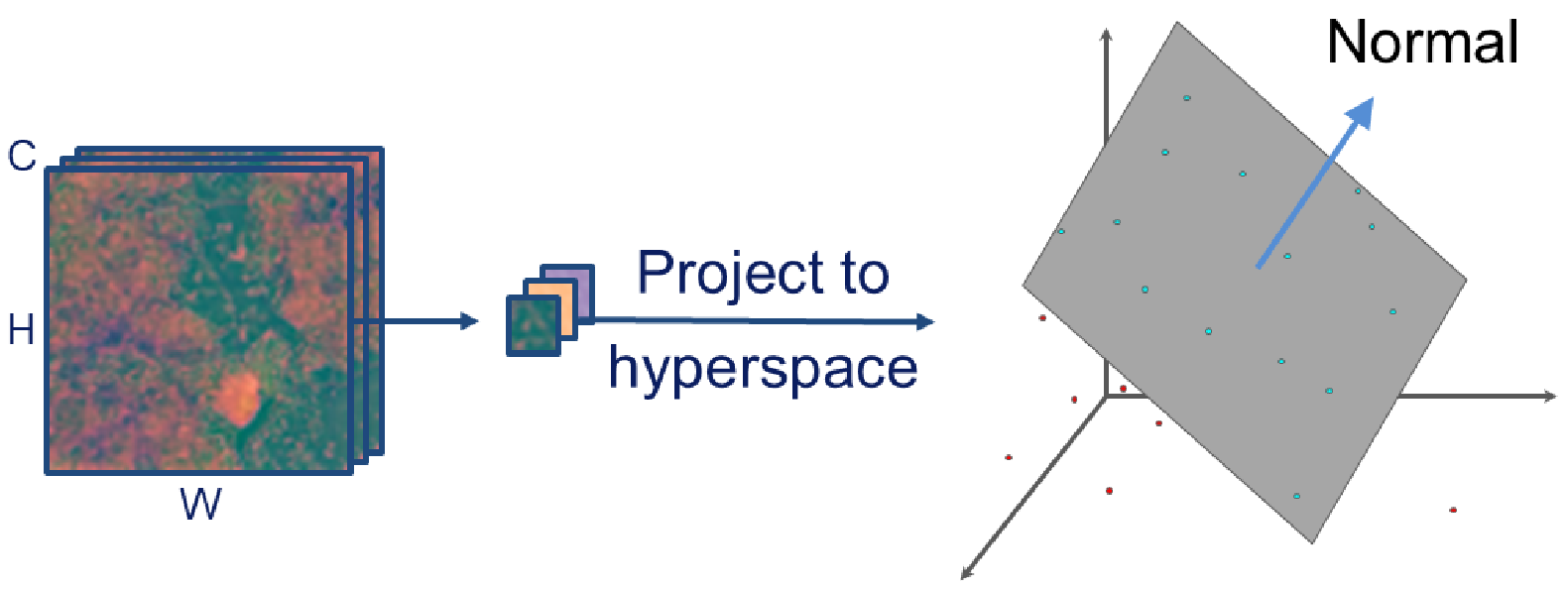}
    \caption{Global Concept Tokens: Input feature maps are processed by a linear layer followed by a Softmax function to generate conceptual attention maps. Each feature is a treated as a semantic vector projected onto a hyperplane.}
    \label{fig:concept}
\end{figure}

\section{Conceptual Attention Transformation}\label{subsec:conceptual-attention-transformation} Kernels in the convolutional pooling layer increasingly limit spatially-distant interaction. We complement this effect with an attention-based module to enhance global feature interaction.  We aim to leverage spatial attention to transform feature maps into a compact set of semantic tokens. Although our conceptual attention transformation (Figure \ref{fig:conceptual-attention-transformation}) is inspired by the high-level conceptual features introduced in the Visual Transformer architecture ~\cite{Wu2021VisualTransformer}, several novel contributions make our approach unique.  A clear departure from prior work is a novel projection layer that integrates the input with the semantic conceptual tokens.

\vspace{10pt}
\noindent Let $X$ represent the input feature map of shape  $(C, H, W)$, where $C$, $H$, and $W$ denote the number of channels, height, and width, respectively. The transformation stage begins by incorporating positional embedding information into the input features $X$ and applying a convolutional operation with a kernel size of $3x3$ for mixing. This operation is expressed in Equation \ref{eq:pos-mixing}:

\begin{equation}\label{eq:pos-mixing}
X_p = Conv(X + PE)
\end{equation}

\vspace{10pt}
\noindent We transform the positionally mixed features $X_p$ into $L$ conceptual representations, which can be learned in either an input-independent or input-dependent manner. For input-independent concepts, a linear projection maps the $C$ features into $L$ concepts, enabling the computation of similarity scores between each feature and each concept. This projection defines a hyperspace in $\mathbb{R}^C$, where the dot product measures the relevance of features to each concept. For input-dependent concepts, the process begins by extracting conceptual representations from the input features. This is achieved through a sequence of convolutional layers with a kernel size of $3x3$, followed by max pooling with a size of $2x2$ and average pooling along the spatial dimensions. This sequence reduces the input feature map into a single feature token of shape $(C, 1)$. The reduced features are then projected into $(C, L)$ using a linear layer, producing $L$ concepts, each represented in $\mathbb{R}^C$. 

\vspace{10pt}
\noindent Concepts (Figure~\ref{fig:concept}) are used to compute similarity scores for each visual feature in the input as shown in Equation \ref{eq:conceptual-attention-unnorm}:
\begin{equation}\label{eq:conceptual-attention-unnorm}
attn_{s} = W_{con}^T X_p
\end{equation}

\vspace{10pt}
\noindent Here, $attn_{s}$ represents the unnormalized conceptual attention transformation, and $W^T_{con} \in \mathbb{R}^{C \times L}$ denotes the conceptual hyperplanes in $\mathbb{R}^C$. The unnormalized attention $attn_{s}$ is passed through a Softmax layer to compute the conceptual attention map $attn \in \mathbb{R}^{L \times HW}$, as defined in Equation \ref{eq:normalize-conceptual-attention}:
\begin{equation}\label{eq:normalize-conceptual-attention}
attn = \text{Softmax}(attn_{s})
\end{equation}

\vspace{10pt}
\noindent This attention map indicates, for each concept $l \in \{1, \ldots, L\}$, an attention matrix of shape $(1, HW)$ that identifies which features of the input $X_p$ are relevant to concept $l$. Using this attention, the features for each concept are computed through attention pooling, as shown in Equation \ref{eq:concept-pooling}:
\begin{equation}\label{eq:concept-pooling}
T_{c} = attn \cdot X_p^{\prime}
\end{equation}

Here, $T_{c} \in \mathbb{R}^{L \times C}$ are the conceptual tokens, and $X_p^{\prime} \in \mathbb{R}^{HW \times C}$ is the reshaped tensor for $X_p$.

 \vspace{10pt}
\noindent To integrate information from the conceptual tokens back into the input features, the conceptual attention map, $attn$, is reused to compute the backward flow contribution of each visual token to the conceptual tokens. For each concept $l \in \{1, \ldots, L\}$, the attention map $attn^{l} \in \mathbb{R}^{HW}$ distributes contributions across $HW$ visual tokens, indicating the degree to which each visual token contributes to a concept. This backward flow is computed as described in Equation \ref{eq:attention-contribution}:
\begin{equation}\label{eq:attention-contribution}
attn_{\mu} = W \cdot (attn + \alpha)  
\end{equation}
The stochasticity term $\alpha$  introduces  variance into the backward flow to ensure  tokens are updated with sufficient diversity to mitigate smoothing that occurs when tokens become increasingly similar after each update. This occurs when dot products incorrectly map highly similar features or features that exhibit similar backward contributions to the same concept.  We can learn $\alpha$ independent of input features using positional information as a positional bias. It can also be learned in a feature-dependent manor by applying aggressive convolutional pooling to compute a parameter of shape $(H, W, C)$. This parameter is projected via  convolutional to produce $L$ channels from $C$ channels, yielding $\alpha$. 

\vspace{10pt}
\noindent Using the contribution attention $attn_{\mu}$, we multiply this value with the conceptual tokens $T_c$ to compute the mixed value defined in Equation \ref{eq:compute-mixed-features}:
\begin{equation}\label{eq:compute-mixed-features}
\phi = \text{GELU}(W_{m}(attn_{\mu} \cdot T_{c}))
\end{equation}

\noindent The idea is to compute a mixed term $\phi$  that can be used to aggregate with the input value $X$. The final update step is performed by combining the linearly projected input with $\phi$ as shown in the Equation \ref{eq:update-input-with-mixed}.
\begin{equation}\label{eq:update-input-with-mixed}
X^{g} = W_{o} \cdot X + \phi
\end{equation}

\noindent The resulting output, $X^{g}$, is encoded with the globally informed parameter $\phi$, enabling the original input $X$ to achieve global contextual interactions with other feature tokens.

\section{Experimental Setup} \label{sec:experiments}

\vspace{10pt}
\noindent To demonstrate the versatility of our enhanced architecture, we  evaluated its performance for object detection tasks using three transformer object detection frameworks: the standard Vision Transformer (ViT)~\cite{10.1007/978-3-031-20077-9_17}, the Swin Transformer ~\cite{9710580}, and the Deformable Attention Transformer (DAT++) ~\cite{DBLP:journals/corr/abs-2201-00520,xia2023datspatiallydynamicvision}. These architectures represent a diverse set of self-attention mechanisms, ranging from standard self-attention to advanced techniques such as shifted window attention and deformable attention. Our analysis includes several benchmark datasets, with a particular emphasis on medical datasets including one new dataset of cancerous tumors (chimeric cell clusters) which we contribute to the research community. The results demonstrate that our module can seamlessly integrate with diverse transformer architectures and significantly enhance their overall performance, particularly on complex dataset domains. We discuss the implementation details of our interaction enhancement modules for different vision transformer models in Section \ref{subsec:model-setup}. The datasets and training procedure are provided in Section \ref{subsec:datasets}.

\begin{figure*}[t]
    \centering
    \begin{subfigure}{0.18\textwidth} 
        \centering
        \includegraphics[width=3cm, height=3cm]{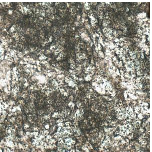}
        \caption{CCellBio }
        \label{fig:image1}
    \end{subfigure}%
    \hspace{-0.01\textwidth} 
    \begin{subfigure}{0.18\textwidth} 
        \centering
        \includegraphics[width=3cm, height=3cm]{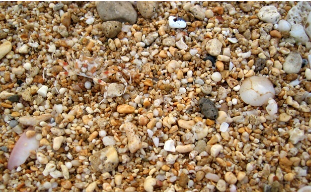}
        \caption{COD10K-V2 }
        \label{fig:image2}
    \end{subfigure}%
    \hspace{-0.01\textwidth} 
    \begin{subfigure}{0.18\textwidth} 
        \centering
        \includegraphics[width=3cm, height=3cm]{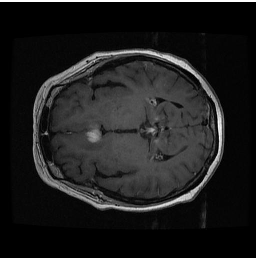}
        \caption{Brain Tumor }
        \label{fig:image3}
    \end{subfigure}%
    \hspace{-0.01\textwidth} 
    \begin{subfigure}{0.18\textwidth} 
        \centering
        \includegraphics[width=3cm, height=3cm]{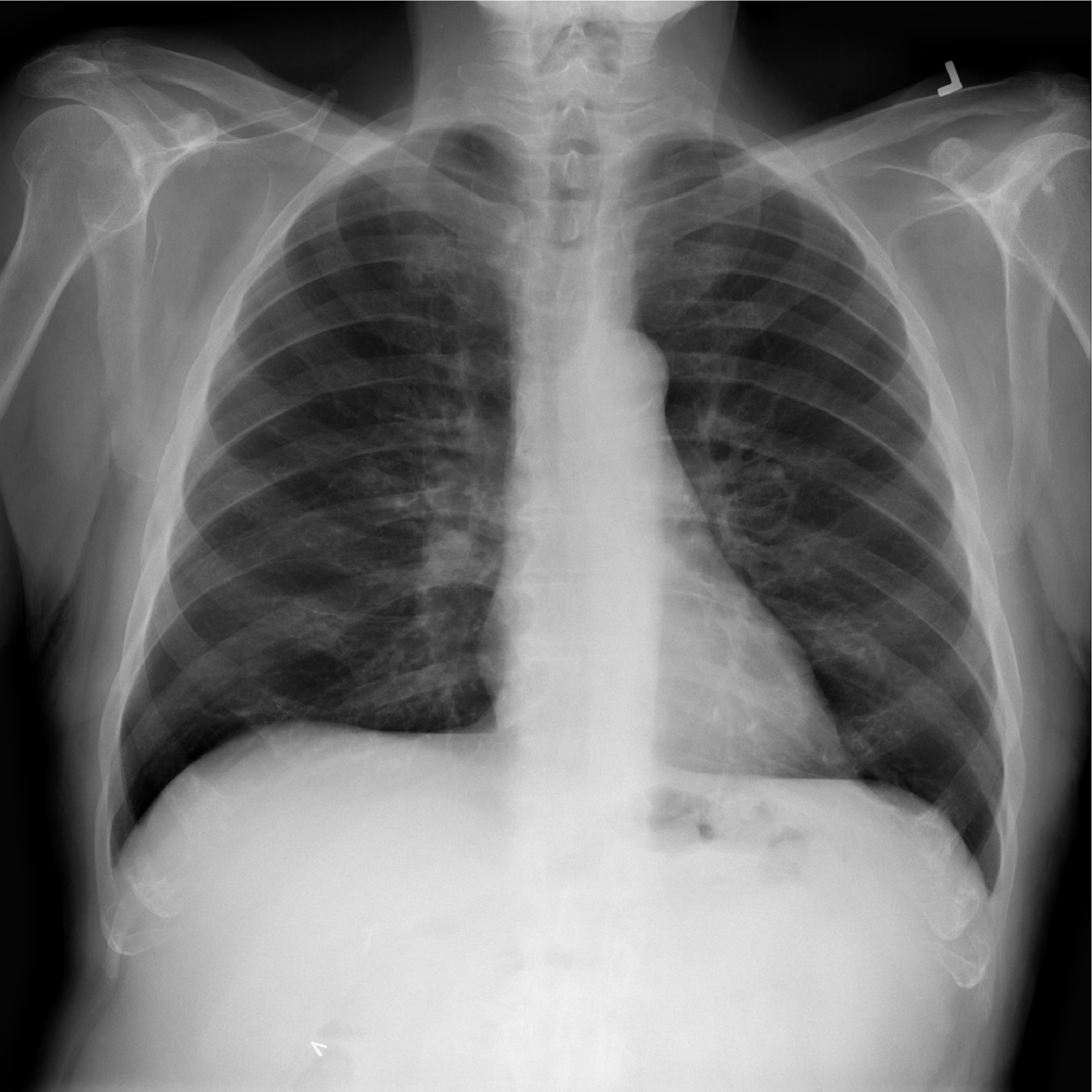}
        \caption{NIH Chest XRay }
        \label{fig:image4}
    \end{subfigure}%
    \hspace{-0.01\textwidth} 
    \begin{subfigure}{0.18\textwidth} 
        \centering
        \includegraphics[width=3cm, height=3cm]{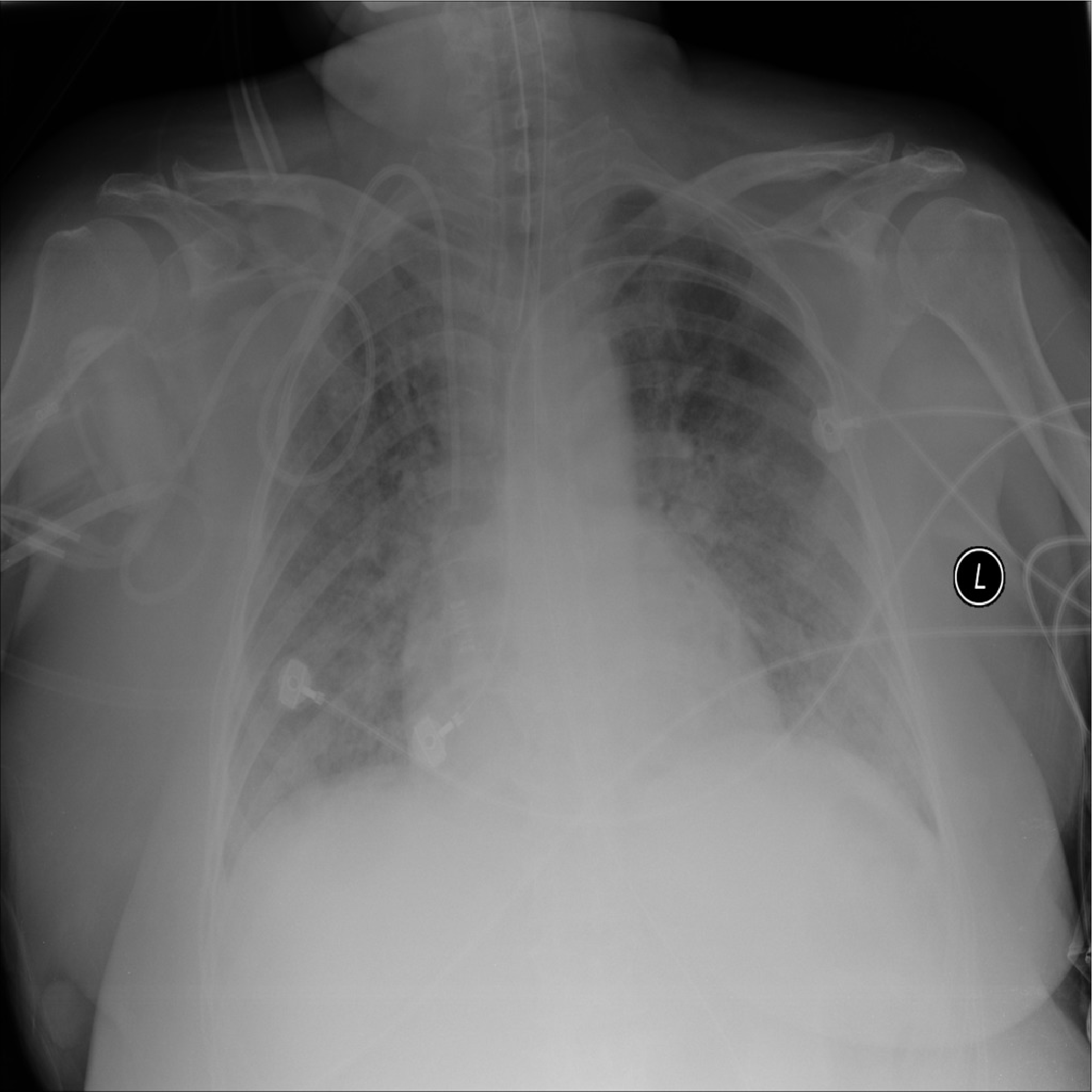}
        \caption{RSNA Pneumonia}
        \label{fig:image5}
    \end{subfigure}
    \caption{Examples of each dataset are as follows: (a) CCellBio Dataset: This dataset contains $3,643$ test images and $26,991$ training images, all showing cancer tumors in stained tissue scans. (b) COD10K-V2: This dataset contains $4,000$ test images and $6,000$ training images and focuses on concealed objects in both natural and artificial environments, with an example image displaying a camouflaged crab in the top-left quarter. (c) Brain Tumor Dataset: This medical dataset includes T1-weighted contrast-enhanced images, featuring three types of brain tumors: meningioma, glioma, and pituitary tumor with $613$ test images and $2,451$ training images. (d) NIH Chest XRay Dataset containing XRay Chest imaging with $1,000$ bounding box annotations with $176$ test images and $704$ training images. (e) RSNA Pneumonia Dataset to detect lung opacity with $1,911$ testing, $7,644$.}
    \label{fig:datasets}
\end{figure*}

\subsection{Datasets}\label{subsec:datasets} Figure \ref{fig:datasets}  shows the range of benchmark datasets for our evaluation. For each dataset, we minimize architectural modifications, only adjusting the patch embedding size, input resolution, and the number of output classes according to the provided configurations. Our evaluation includes testing on our new CCellBio dataset, available with this publication, which contains 26,991 training images and 3,643 test images of cancer tumors in tissue scans along with ground truth annotations. In addition, we assess our models on several publicly available datasets: COD10K-V2 ~\cite{DBLP:journals/corr/abs-2102-10274, fan2020camouflaged}, Brain Tumor Detection ~\cite{Cheng2017}, VinDr-CXR ~\cite{nguyen2022vindrcxropendatasetchest, nguyen2022vindrcxropendatasetchest, goldberger2000physionet}, NIH-ChestXRay ~\cite{Wang_2017}, which includes nearly 1,000 images with bounding box annotations for 8 categories, and the RSNA Pneumonia Detection Dataset ~\cite{Wang_2017, Shih2019RSNAPneumonia, NIHChestXray, RSNAPneumonia}, which includes 7,644 training images and 1,911 test images for detecting lung opacity.

\vspace{10pt}
 \noindent We apply data augmentation in a consistent manor across all datasets studied in the following sequence: random flip with a probability of 50\%, random resizing within a ratio range of 0.1 to 2.0 maintaining aspect ratio, and random cropping. Next, we remove annotations with spatial dimensions (height or width) smaller than $10^{-2}$ for more stable training before mean value normalization (123.675, 116.28, 103.53), and standard deviation values of (58.395, 57.12, 57.375) for the three color channels respectively. Finally, padding is applied to match the defined image size (114, 114, 114) for the height, width, and channels respectively.

\subsection{Model Configurations}\label{subsec:model-setup}
Our study centers on improving the transformer backbone architecture, a model used for feature extraction for higher level computer vision tasks; in our case object detection. Thus, we evaluated the aforementioned transformer models (standard Vision ViT~\cite{10.1007/978-3-031-20077-9_17}, Swin~\cite{9710580}, and DAT++ ~\cite{DBLP:journals/corr/abs-2201-00520,xia2023datspatiallydynamicvision})  within the RetinaNet framework ~\cite{DBLP:journals/corr/abs-1708-02002},  a single, unified network composed of a backbone network and two task-specific subnetworks. Nevertheless, the proposed approach is versatile and can be extended to other detection frameworks. We now describe a comprehensive evaluation approach that underscores the adaptability of our module across diverse transformer architectures, affirming its potential for broader generalization.

\textbf{Backbone Configurations:} For each transformer backbone, we derive two versions: the original architecture  (ViT, Swin, and DAT), and a corresponding modified version incorporating our enhanced interaction modules (EI-ViT, EI-Swin, and EI-DAT). To ensure a fair evaluation, the enhanced architecture configuration is preserved to match that of the original architecture, and we increased the hidden dimensions of the baseline model to approximate the parameter count of the enhanced interaction architecture. We maintained consistent configurations for all architectural components substituting only the backbone to benchmark their relative performances. For both ACP and CAT, the convolution dimensions and the number of concepts are set to match the hidden dimension of each layer in the baseline backbones. Similarly, the number of heads in the Conceptual Attention Transformer matches the number of attention heads in the baseline.

\begin{figure}[t]
    \includegraphics[width=\columnwidth]{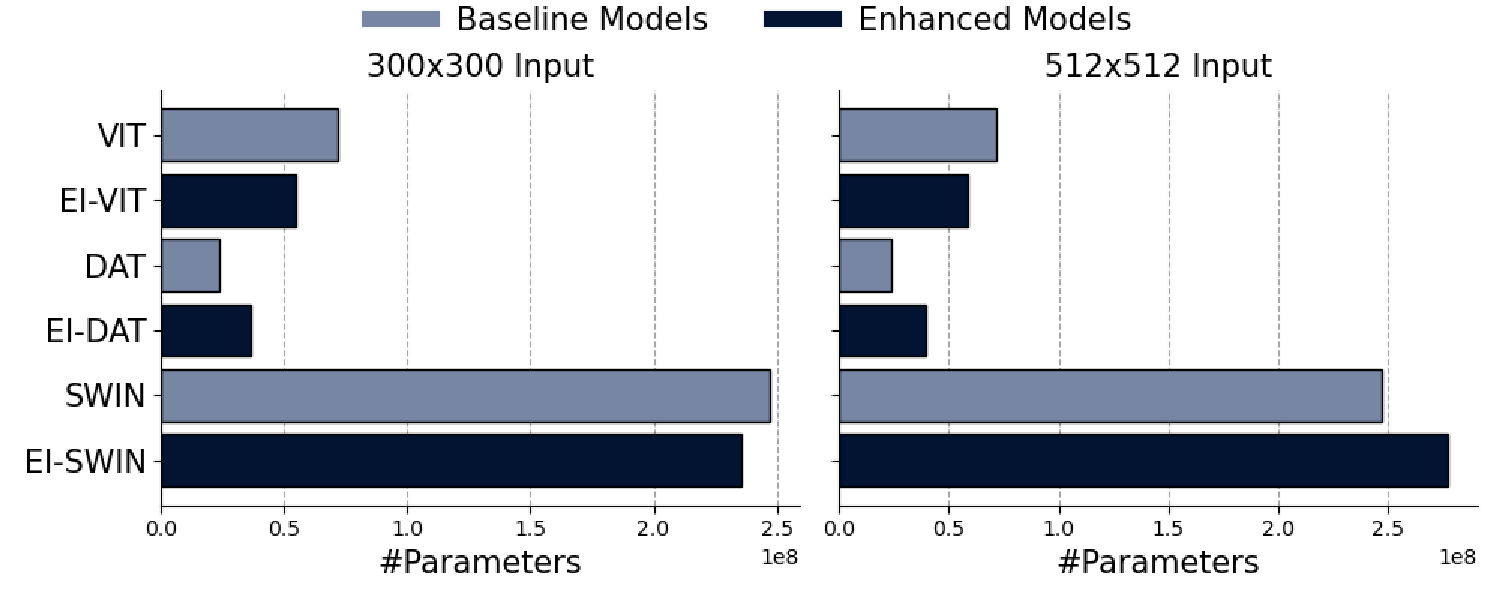}
    \caption{Number of parameters for baseline and enhanced interaction models for 300x300 image input (left) and 512x512 image input (right).}
    \label{fig:params}
\end{figure}

\begin{table}[b]
\centering
\caption{Parameter Configuration: Bounding Box Head}
\label{table:bbox_head_configuration} \label{tab:rentinahead-config}
\begin{tabular}{ll}
\toprule
\textbf{Attribute}                   & \textbf{Value}                     \\ 
\midrule
Name                                 & RetinaHead                         \\ 
Number of Classes                    & 3                                  \\ 
Input Channels                       & 256                                \\ 
Stacked Convolution Layers           & 4                                  \\ 
Feature Channels                     & 256                                \\ 
\midrule
\textbf{Anchor Generator}            &                                     \\ 
\quad Name                           & Anchor Generator                   \\ 
\quad Octave Base Scale              & 4                                  \\ 
\quad Scales Per Octave              & 3                                  \\ 
\quad Aspect Ratios                  & 0.5, 1.0, 2.0                     \\ 
\quad Anchor Strides                 & 8, 16, 32, 64, 128                \\ 
\midrule
\textbf{Bounding Box Coder}          &                                     \\ 
\quad Name                           & DeltaXYWHBBoxCoder                 \\ 
\quad Target Means                   & 0.0, 0.0, 0.0, 0.0                \\ 
\quad Target Stds     & 1.0, 1.0, 1.0, 1.0                \\ 
\midrule
\textbf{Classification Loss}         &                                     \\ 
\quad Name                           & Focal Loss                         \\ 
\quad Use Sigmoid                    & True                               \\ 
\quad Gamma                          & 2.0                                \\ 
\quad Alpha                          & 0.25                               \\ 
\quad Loss Weight                    & 1.0                                \\ 
\midrule
\textbf{Bounding Box Loss}           &                                     \\ 
\quad Name                           & L1 Loss                            \\ 
\quad Loss Weight                    & 1.0                                \\ 
\bottomrule
\end{tabular}
\end{table}
\noindent

Our enhanced interaction methods primarily increase channel width rather than the number of transformer blocks. Thus we scale the baseline model width to approximate parameter counts for enhanced models to facilitate a fair comparison in feature and attention map analysis. We ensured that both models maintain identical depth and number of attention heads. We categorized the baseline models into two groups based on the spatial dimensions of the input as shown in Figure~\ref{fig:params}.  For the CCellBio dataset with $300 \times 300$ input images, the number of parameters for the baseline VIT, DAT, and SWIN transformers are $71.6$M, $23.5$M, and $247.1$M, respectively, while enhanced versions are $54.5$M, $36.2$M, and $247.1$M parameters receptively. For the COD10k-V3, NIH Chest XRay, and RSNA Pneumonia datasets with $512 \times 512$ input images, we scaled the number of parameters for the baseline VIT, DAT, and SWIN to $71.6$M, $23.9$M, and $247.1$M, respectively, compared to the enhanced models, which have $58.7$M, $39.6$M, and $277.1$M parameters. 
 
 \begin{table}[h] 
\centering
\caption{Parameter Configuration: for FPN}
\label{table:neck_configuration} \label{tab:fpn-config}
\begin{tabular}{ll}
\toprule
\textbf{Attribute}          & \textbf{Value}                        \\ 
\midrule
Name                        & FPN                                             \\ 
Output Dim                  & 256                                   \\ 
Levels                      & Start: 1, Out: 5                      \\ 
\bottomrule
\end{tabular}
\end{table}

\vspace{10pt}
\noindent This approach minimizes architectural modifications, allowing us to attribute performance gains to improved self-attention interactions rather than increased model size.  Our implementation uses a Feature Pyramid Network (FPN) ~\cite{DBLP:journals/corr/LinDGHHB16} to enhance hierarchical feature extraction across multiple layers, a common characteristic of transformer backbones. The FPN output is then processed by the RetinaNet head ~\cite{DBLP:journals/corr/abs-1708-02002} to predict bounding boxes. Detailed configurations for each component are summarized in Tables~\ref{tab:fpn-config} and~\ref{tab:rentinahead-config}. All backbone configurations for the baseline and their enhanced interactive architectures are provided in the Appendix.

\textbf{Training and Testing:} The baseline models and their enhanced interaction architectures were trained independently on benchmark datasets for 30 epochs, with evaluation conducted on the respective test sets. Training was carried out using randomly initialized weights, emphasizing the fact that the proposed enhanced interaction modules do not necessitate pre-training. We evaluate mean average precision (mAP) at different IoU thresholds (mAP50 and mAP75) and average recall (AR).

\section{Results \& Analysis}

Here we present results of our benchmark evaluations for the CCellBio, COD10K-V2, and Brain Tumor and RSNA Pneumonia datasets (Tables~\ref{table:CCellBio_model_results},~\ref{table:cod10k_model_results},~\ref{table:braintumor_model_results}, and ~\ref{table:rsna_model_results} respectively). We begin with a summary of the findings discussed in this section. ACP and CAP enhanced interaction components:

\paragraph{Improve both mAP and AR metrics across five challenging detection datasets} (Tables~\ref{table:CCellBio_model_results},~\ref{table:cod10k_model_results},~\ref{table:braintumor_model_results},~\ref{table:nihchestxray_model_results}, and~\ref{table:rsna_model_results}) and are more effective in improving mAP than AR (Figures~\ref{fig:map-cross-dataset} and~\ref{fig:ar-cross-dataset}). 

\paragraph{Improve feature representation and reduce over-smoothing} by allowing features to interact both locally and globally evidenced by sharper feature maps that  discriminate visually similar objects more effectively (Figure~\ref{fig:feature-analysis-pca}). 

\paragraph{Alter attention behaviors prior to self-attention}, where the network demonstrates reduced attention in earlier blocks and increased attention activity in later blocks with more descriptive semantic representations (Figure~\ref{fig:attention-map-analysis}). 

\paragraph{Improve performance over baseline models without relying on an increased number of parameters or additional modules} (Figures~\ref{fig:kernel-cka}).

\subsection{Quantitative Benchmark Dataset Analysis}

\vspace{10pt}
\noindent {CCellBio:} The evaluation on the CCellBio dataset, which focuses on cancer tumor detection in stained tissue scans, demonstrates consistent improvements with the enhanced interaction architectures across most metrics. EI-VIT outperformed the baseline with a 9.14\% increase in mAP, a 5.69\% improvement in mAP50, a 14.58\% improvement in mAP75, and a 5.71\% improvement in AR. Similarly, EI-DAT achieved a 1.92\% increase in mAP, a notable 3.49\% improvement in mAP50, and a 1.50\% improvement in AR. EI-SWIN consistently outperformed the baseline across all metrics. It demonstrated a 7.85\% increase in mAP, a 6.21\% improvement in mAP50, a 14.65\% increase in mAP75, and a 5.50\% improvement in AR. These results confirm the generalizability and effectiveness of the proposed interaction modules across different backbone architectures and underscore their potential for improving cancer tumor detection in medical imaging applications.

\vspace{10pt}
\noindent {COD10K-V3:} Enhanced interaction architectures consistently outperformed their respective baselines across all metrics on the COD10K-V3 dataset. EI-VIT achieved a significant improvement with a 21.05\% increase in mAP, a 17.45\% increase in mAP50, and a 166.67\% increase in mAP75. However, there was a 0.36\% decline in relative performance compared to the baseline VIT model due to challenges processing deformable points which we discuss in Section~\ref{sec:limit}.  EI-DAT showed a 15.15\% improvement in mAP, a 9.66\% improvement in mAP50, a 42.86\% increase in mAP75, and a 2.99\% gain in AR, AR300, and AR1000. \textbf{The most notable performance gain was observed with EI-SWIN} which achieved a 103.03\% improvement in mAP, a 73.85\% increase in mAP50, a 375.00\% gain in mAP75, and a 24.91\% increase across AR. Thus we improve AR and precision at higher IoU thresholds (mAP75), which are crucial for detecting concealed objects in challenging scenarios such as those in the COD10K-V2.

\vspace{10pt}
\noindent\emph{Brain Tumor:} The enhanced architectures outperform the baseline on all mAP and AR metrics for brain tumor detection as shown in Table \ref{table:braintumor_model_results}. Specifically, ViT shows relative performance gains of 21.73\%, 13.93\%, 37.07\%, and 7.84\% for mAP, mAP50, mAP75, and AR, respectively. The DAT model also shows positive improvements with 2.71\%, 1.8\%, 5.68\%, and 0.76\% gains. The Swin Transformer demonstrates the largest improvements, with gains of 63.92\%, 46.64\%, 87.13\%, and 15.65\% for mAP, mAP50, mAP75, and AR, respectively.

\begin{table}[t]
\centering
\caption{Model Evaluation on CCellBio Dataset}
\label{table:CCellBio_model_results}
\begin{tabular}{l@{\hskip 7pt}c@{\hskip 7pt}c@{\hskip 7pt}c@{\hskip 7pt}c}
\toprule
\textbf{Model} & \textbf{mAP} & \textbf{mAP50} & \textbf{mAP75} & \textbf{AR} \\
\midrule
VIT & 0.350 & 0.738 & 0.288 & 0.473 \\
EIVIT & \textbf{0.382} & \textbf{0.780} & \textbf{0.330} & \textbf{0.500} \\
& \hphantom{0}\textcolor{darkgreen}{+9.14\%\hphantom{0}} & \hphantom{0}\textcolor{darkgreen}{+5.69\%}\hphantom{0} & \textcolor{darkgreen}{+14.58\%} & \hphantom{0}\textcolor{darkgreen}{+5.71\%}\hphantom{0} \\
\midrule
DAT & 0.417 & 0.831 & 0.357 & 0.534 \\
EIDAT & \textbf{0.425} & \textbf{0.860} & \textbf{0.355} & \textbf{0.542} \\
& \textcolor{darkgreen}{+1.92\%} & \textcolor{darkgreen}{+3.49\%} & \textcolor{red}{-0.56\%} & \textcolor{darkgreen}{+1.50\%} \\
\midrule
SWIN & 0.382 & 0.789 & 0.314 & 0.509 \\
EISWIN & \textbf{0.412} & \textbf{0.838} & \textbf{0.360} & \textbf{0.537} \\
& \textcolor{darkgreen}{+7.85\%} & \textcolor{darkgreen}{+6.21\%} & \textcolor{darkgreen}{+14.65\%} & \textcolor{darkgreen}{+5.50\%} \\
\bottomrule
\end{tabular}
\end{table}

\begin{table}[h!]
\centering
\caption{Model Evaluation on COD10K-V2 Dataset}
\label{table:cod10k_model_results}
\begin{tabular}{l@{\hskip 7pt}c@{\hskip 7pt}c@{\hskip 7pt}c@{\hskip 7pt}c}
\toprule
\textbf{Model} & \textbf{mAP} & \textbf{mAP50} & \textbf{mAP75} & \textbf{AR} \\
\midrule
VIT & 0.038 & 0.149 & 0.003 & 0.279 \\
EIVIT & \textbf{0.046} & \textbf{0.175} & \textbf{0.008} & \textbf{0.278} \\
& \textcolor{darkgreen}{+21.05\%} & \textcolor{darkgreen}{+17.45\%} & \textcolor{darkgreen}{+166.67\%} & \textcolor{red}{-0.36\%} \\
\midrule
DAT & 0.066 & 0.238 & 0.014 & 0.335 \\
EIDAT & \textbf{0.076} & \textbf{0.261} & \textbf{0.020} & \textbf{0.345} \\
& \textcolor{darkgreen}{+15.15\%} & \textcolor{darkgreen}{+9.66\%} & \textcolor{darkgreen}{+42.86\%} & \textcolor{darkgreen}{+2.99\%} \\
\midrule
SWIN & 0.033 & 0.130 & 0.004 & 0.269 \\
EISWIN & \textbf{0.067} & \textbf{0.226} & \textbf{0.019} & \textbf{0.336} \\
& \textcolor{darkgreen}{+103.03\%} & \textcolor{darkgreen}{+\hphantom{0}73.85\%} & \textcolor{darkgreen}{+375.00\%} & \textcolor{darkgreen}{+24.91\%} \\
\bottomrule
\end{tabular}
\end{table}

\vspace{10pt}
\paragraph{NIH Chest XRay:} Table \ref{table:nihchestxray_model_results} presents the model evaluation results on the NIH Chest XRay Detection dataset.  EI-ViT outperforms the standard ViT model on mAP, mAP75, and AR with 7.69\%, 40\%, and 3.91\% improvement accordingly. However, there is 21.95\% decrease in mAP75 metric. The relatively small size of the dataset may explain the lower performance scores observed when models are not initialized with pretrained weights, as limited data can hinder the model's ability to generalize effectively (Section~\ref{sec:limit}). EI-DAT shows significant gains over the plain DAT model, with increases of 30.77\% in bbox\_mAP, 40.00\% in bbox\_mAP\_50, and 27.27\% in bbox\_mAP\_75, particularly improving performance in bbox\_mAP\_50.  EI-Swin also shows improvements over the plain Swin Transformer, with increases of 121.43\% in bbox\_mAP, 102.50\% in bbox\_mAP\_50, and 180.00\% in AR. Enhanced interaction modules consistently improves detection performance across the transformer models. However, for small datasets like NIH Chest XRay  (see Section~\ref{sec:limit}), without pretrained weights, prediction rates are lower, a challenge that affects model generalization to small datasets.

\vspace{10pt}
 \paragraph{RSNA Pneumonia:} For the RSNA Pneumonia dataset, Table \ref{table:rsna_model_results}, EI-VIT, EI-DAT and EI-SWIN consistently demonstrate superior performance compared to their baseline counterparts. Relative improvements vary across models and benchmarks, highlighting the effectiveness of enhanced contextual interactions in specific architectures. EI-VIT achieves a 7.27\% improvement in \textit{bbox\_mAP}, a 4.39\% increase in \textit{bbox\_mAP\_50}, and a remarkable 14.29\% boost in \textit{bbox\_mAP\_75}, showcasing substantial gains in both coarse and fine-grain detection accuracy. There is a marginal improvement for recall of 0.43\% across all \textit{bbox\_AR} metrics, indicating consistent object detoxification performance under varying IoU thresholds. EI-DAT presents mixed results which we explain in Section~\ref{sec:limit}, while \textit{bbox\_AR} metrics improve by 0.64\% for better recall. The \textit{bbox\_mAP} has a slight decrease of 1.69\%, and \textit{bbox\_mAP\_75} drops by 4.55\%. However, EI-DAT still achieves a modest 1.15\% gain in \textit{bbox\_mAP\_50}, indicating that its enhancements are more effective for lower IoU thresholds. These results suggest potential limitations in fine-grained detection for  EI-DAT. EI-SWIN delivers the most significant relative improvements; a 25.53\% increase in \textit{bbox\_mAP}, a 18.67\% gain in \textit{bbox\_mAP\_50}, and an impressive 78.26\% boost in \textit{bbox\_mAP\_75}. Recall that AR improved by 3.56\%, highlighting consistent gains across the board. SWIN's hierarchical architecture is particularly well-suited to benefit from enhanced contextual interactions.

\begin{figure*}[t]
    \centering
    \begin{minipage}{0.45\textwidth}
        \centering
      \includegraphics[width=\columnwidth]{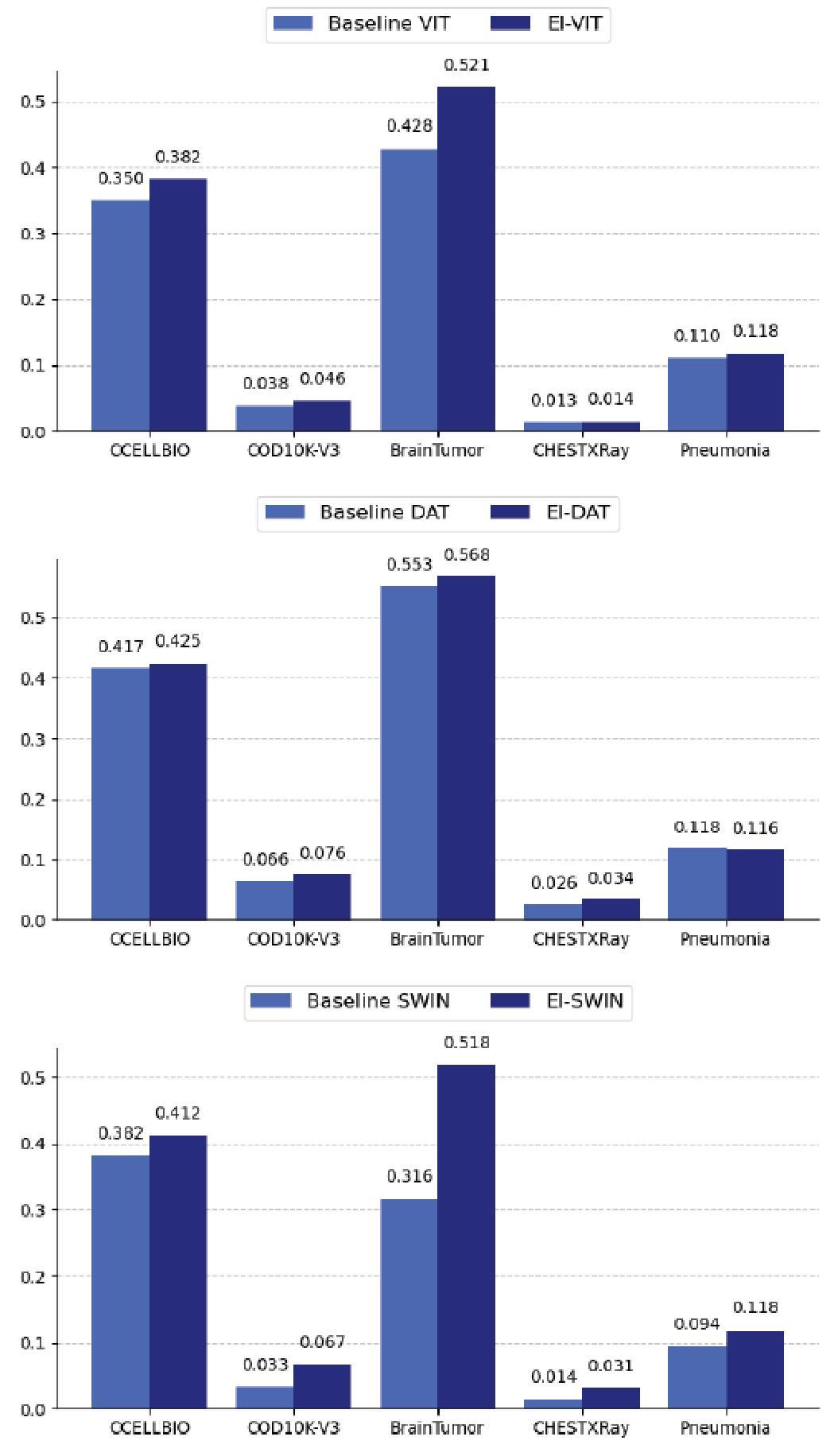}
    \caption{BBox Mean Average Precision (mAP) comparison between the baseline VIT, DAT, and SWIN transformers with their enhanced architectures across 5 datasets.}
   \label{fig:ar-cross-dataset}
    \end{minipage}\hfill
    \begin{minipage}{0.45\textwidth}
        \centering
         \includegraphics[width=\columnwidth]{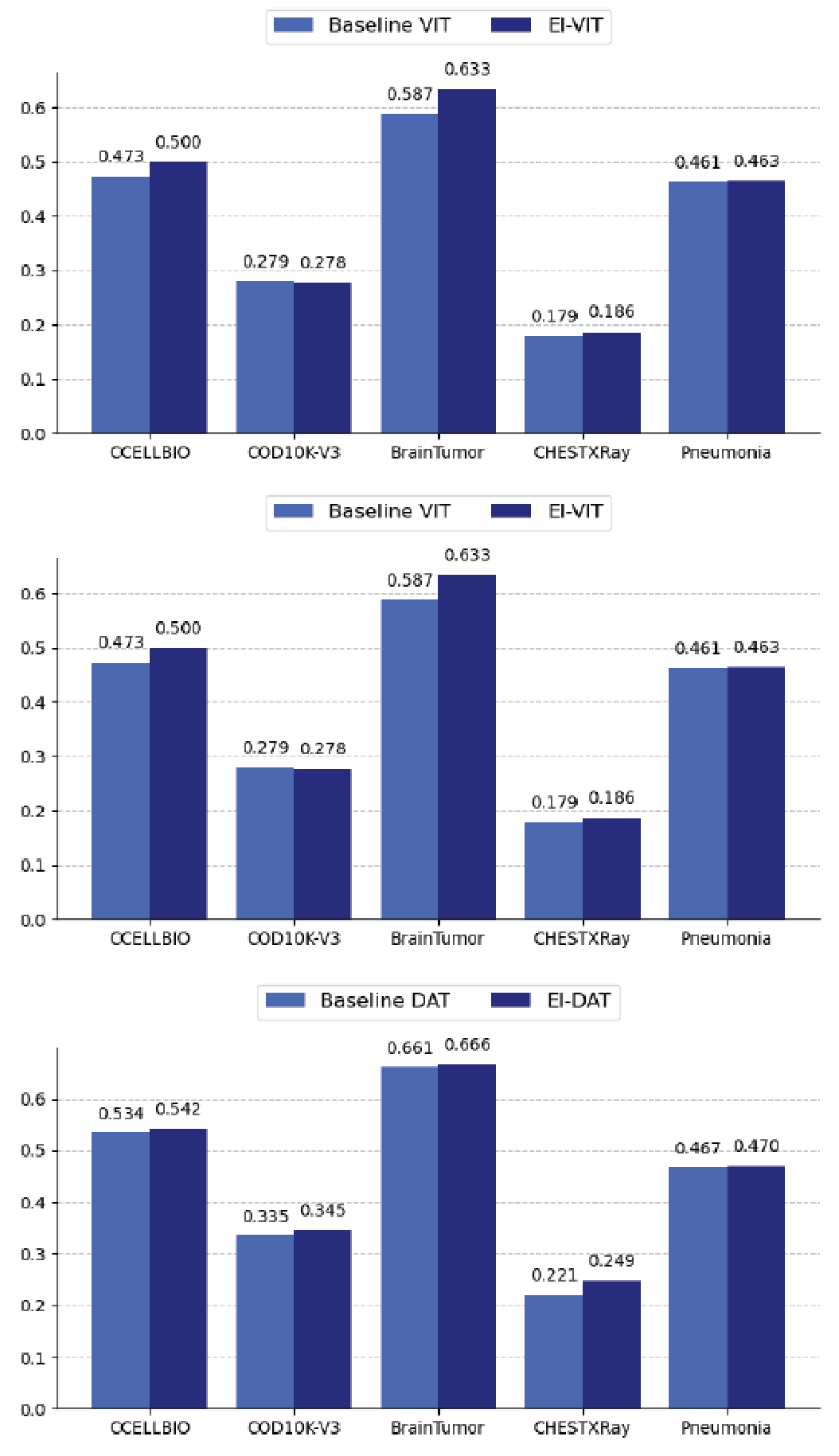}
   \caption{ BBox Average Recall (AR) comparison between the baseline VIT, DAT, and SWIN transformers with their enhanced architectures across 5 datasets.}
    \label{fig:map-cross-dataset}
    \end{minipage}
\end{figure*}

\vspace{10pt}
\noindent \textbf{Cross Dataset Analysis}. Our enhanced interaction modules demonstrate improved mAP metrics across five datasets, as shown in Figure \ref{fig:map-cross-dataset}. Enhanced architectures consistently achieve higher mAP scores, with the Swin Transformer showing the highest average improvement of 6.14\%, followed by ViT (2.84\%) and DAT (0.78\%). The inclusion of enhanced interaction components also improved average recall metrics, as shown in Figure \ref{fig:ar-cross-dataset}. Overall, the relative average AR improvements were 1.62\% for ViT, 1.08\% for DAT, and 4.42\% for Swin Transformers across all five datasets. In terms of maximum relative improvement, the Swin Transformer achieved the highest at 24.91\%, followed by DAT at 12.67\% and ViT at 7.84\%.  In most datasets, incorporating enhanced interactions before multi-head self-attention resulted in consistent improvements, except for the DAT model on the COD10k-V3 dataset, which showed a minor relative drop of 0.36\% which we discuss later in the text. 

Our study reveals that the Swin Transformer benefits the most from enhanced interactions, achieving the highest relative improvements in both AR and mAP metrics. We hypothesize that interactions prior to self-attention refine the feature representation within each subwindow of the Swin Transformer. By adding local and global information to features, these interactions enable the shifted window mechanism to more effectively capture relationships within the enriched window patches.

\begin{table}[t]
\centering
\caption{Model Evaluation on Braintumor Dataset}
\label{table:braintumor_model_results}
\begin{tabular}{l@{\hskip 7pt}c@{\hskip 7pt}c@{\hskip 7pt}c@{\hskip 7pt}c}
\toprule
\textbf{Model} & \textbf{mAP} & \textbf{mAP50} & \textbf{mAP75} & \textbf{AR} \\
\midrule
VIT & 0.428 & 0.776 & 0.410 & 0.587 \\
EIVIT & \textbf{0.521} & \textbf{0.884} & \textbf{0.562} & \textbf{0.633} \\
& \hphantom{0}\textcolor{darkgreen}{+21.73\%} & \textcolor{darkgreen}{+13.92\%} & \hphantom{0}\textcolor{darkgreen}{+37.07\%} & \textcolor{darkgreen}{+7.84\%} \\
\midrule
DAT & 0.553 & 0.887 & 0.599 & 0.661 \\
EIDAT & \textbf{0.568} & \textbf{0.903} & \textbf{0.633} & \textbf{0.666} \\
& \textcolor{darkgreen}{+2.71\%} & \textcolor{darkgreen}{+1.80\%} & \textcolor{darkgreen}{+5.68\%} & \textcolor{darkgreen}{+0.76\%} \\
\midrule
SWIN & 0.316 & 0.581 & 0.303 & 0.556 \\
EISWIN & \textbf{0.518} & \textbf{0.852} & \textbf{0.567} & \textbf{0.643} \\
& \textcolor{darkgreen}{+63.92\%} & \textcolor{darkgreen}{+46.64\%} & \textcolor{darkgreen}{+87.13\%} & \textcolor{darkgreen}{+15.65\%} \\
\bottomrule
\end{tabular}
\end{table}

\begin{table}[h!]
\centering
\caption{Model Evaluation on NIHChestXRay Dataset}
\label{table:nihchestxray_model_results}
\begin{tabular}{l@{\hskip 7pt}c@{\hskip 7pt}c@{\hskip 7pt}c@{\hskip 7pt}c}
\toprule
\textbf{Model} & \textbf{mAP} & \textbf{mAP50} & \textbf{mAP75} & \textbf{AR} \\
\midrule
VIT & 0.013 & 0.041 & 0.005 & 0.179 \\
EIVIT & \textbf{0.014} & \textbf{0.032} & \textbf{0.007} & \textbf{0.186} \\
& \textcolor{darkgreen}{+7.69\%} & \textcolor{red}{-21.95\%} & \textcolor{darkgreen}{+40.00\%} & \textcolor{darkgreen}{+3.91\%} \\
\midrule
DAT & 0.026 & 0.060 & 0.022 & 0.221 \\
EIDAT & \textbf{0.034} & \textbf{0.084} & \textbf{0.028} & \textbf{0.249} \\
& \textcolor{darkgreen}{+30.77\%} & \textcolor{darkgreen}{+40.00\%} & \textcolor{darkgreen}{+27.27\%} & \textcolor{darkgreen}{+12.67\%} \\
\midrule
SWIN & 0.014 & 0.040 & 0.005 & 0.181 \\
EISWIN & \textbf{0.031} & \textbf{0.081} & \textbf{0.014} & \textbf{0.204} \\
& \textcolor{darkgreen}{+121.43\%} & \textcolor{darkgreen}{+102.50\%} & \textcolor{darkgreen}{+180.00\%} & \textcolor{darkgreen}{+12.71\%} \\
\bottomrule
\end{tabular}
\end{table}

\begin{table}[h!]
\centering
\caption{Model Evaluation on RSNA Pneumonia Dataset}
\label{table:rsna_model_results}
\begin{tabular}{l@{\hskip 7pt}c@{\hskip 7pt}c@{\hskip 7pt}c@{\hskip 7pt}c}
\toprule
\textbf{Model} & \textbf{mAP} & \textbf{mAP50} & \textbf{mAP75} & \textbf{AR} \\
\midrule
VIT & 0.110 & 0.342 & 0.035 & 0.461 \\
EIVIT & \textbf{0.118} & \textbf{0.357} & \textbf{0.040} & \textbf{0.463} \\
& \textcolor{darkgreen}{+7.27\%} & \textcolor{darkgreen}{+4.39\%} & \textcolor{darkgreen}{+14.29\%} & \textcolor{darkgreen}{+0.43\%} \\
\midrule
DAT & 0.118 & 0.347 & 0.044 & 0.467 \\
EIDAT & \textbf{0.116} & \textbf{0.351} & \textbf{0.042} & \textbf{0.470} \\
& \textcolor{red}{-1.69\%} & \textcolor{darkgreen}{+1.15\%} & \textcolor{red}{-4.55\%} & \textcolor{darkgreen}{+0.64\%} \\
\midrule
SWIN & 0.094 & 0.300 & 0.023 & 0.450 \\
EISWIN & \textbf{0.118} & \textbf{0.356} & \textbf{0.041} & \textbf{0.466} \\
& \hphantom{0}\textcolor{darkgreen}{+25.53\%} & \hphantom{0}\textcolor{darkgreen}{+18.67\%} & \textcolor{darkgreen}{+78.26\%} & \hphantom{0}\textcolor{darkgreen}{+3.56\%} \\
\bottomrule
\end{tabular}
\end{table}

\subsection{Qualitative Feature and Attention Analysis:}

\begin{figure*}[htpb]
    \centering
    \begin{tabular}{c c}
        \multirow{2}{*}[1.2cm]{\rotatebox{90}{\textbf{ViT}}} &
        \begin{subfigure}{0.16\textwidth} 
            \centering
            \includegraphics[width=2.5cm, height=2.5cm]{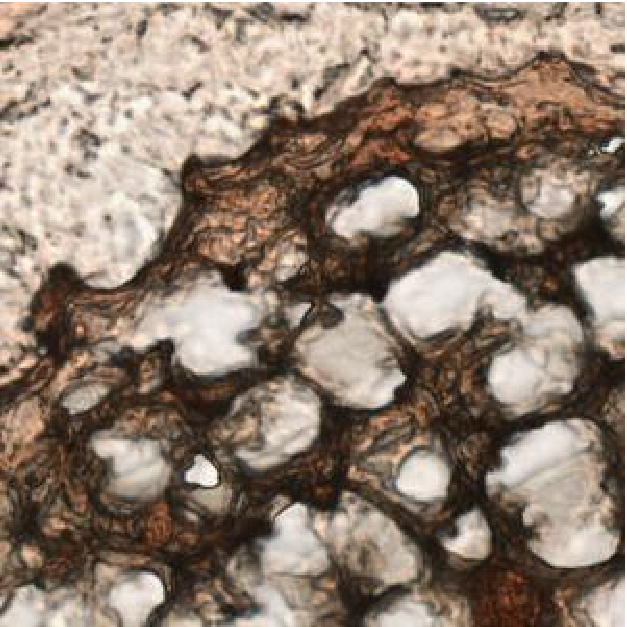}
        \end{subfigure}
        \hspace{-0.02\textwidth}
        \begin{subfigure}{0.16\textwidth} 
            \centering
            \includegraphics[width=2.5cm, height=2.5cm]{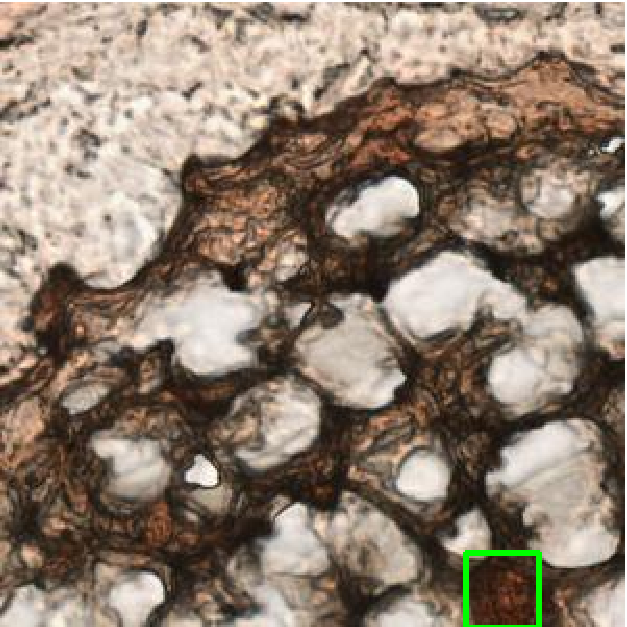}
        \end{subfigure}
        \hspace{-0.02\textwidth}
        \begin{subfigure}{0.16\textwidth} 
            \centering
            \includegraphics[width=2.5cm, height=2.5cm]{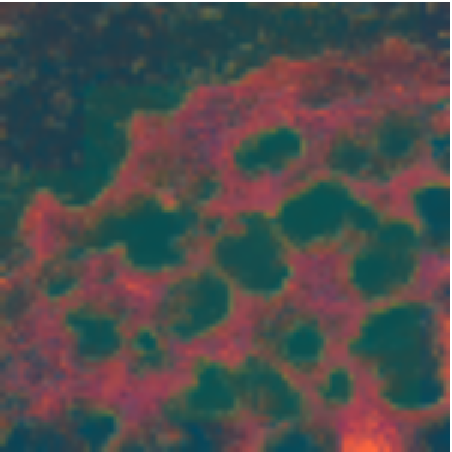}
        \end{subfigure}
        \hspace{-0.02\textwidth}
        \begin{subfigure}{0.16\textwidth} 
            \centering
            \includegraphics[width=2.5cm, height=2.5cm]{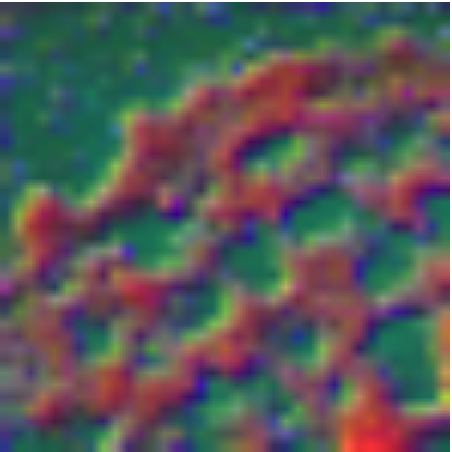}
        \end{subfigure}
        \hspace{-0.02\textwidth}
        \begin{subfigure}{0.16\textwidth} 
            \centering
            \includegraphics[width=2.5cm, height=2.5cm]{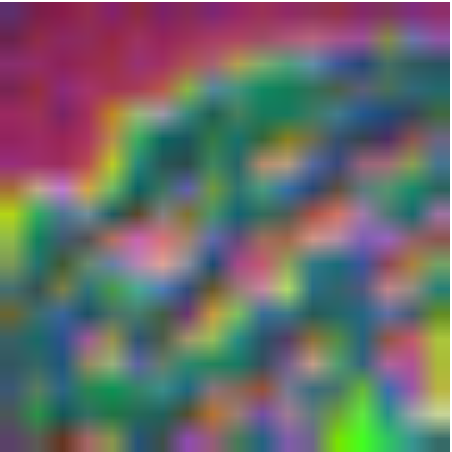}
        \end{subfigure}
        \hspace{-0.02\textwidth}
        \begin{subfigure}{0.16\textwidth} 
            \centering
            \includegraphics[width=2.5cm, height=2.5cm]{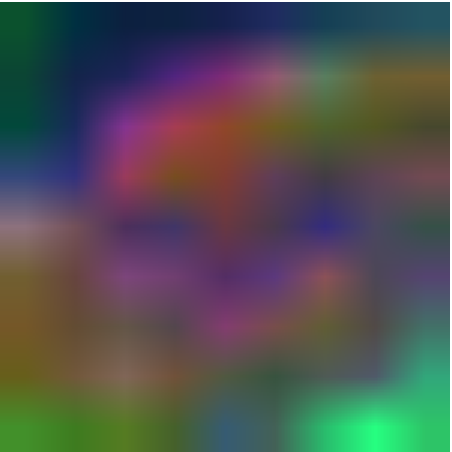}
        \end{subfigure} 
        \\
        \multirow{2}{*}[2.1cm]{\rotatebox{90}{\textbf{EI-ViT}}} &
        \begin{subfigure}{0.16\textwidth} 
            \centering
            \includegraphics[width=2.5cm, height=2.5cm]{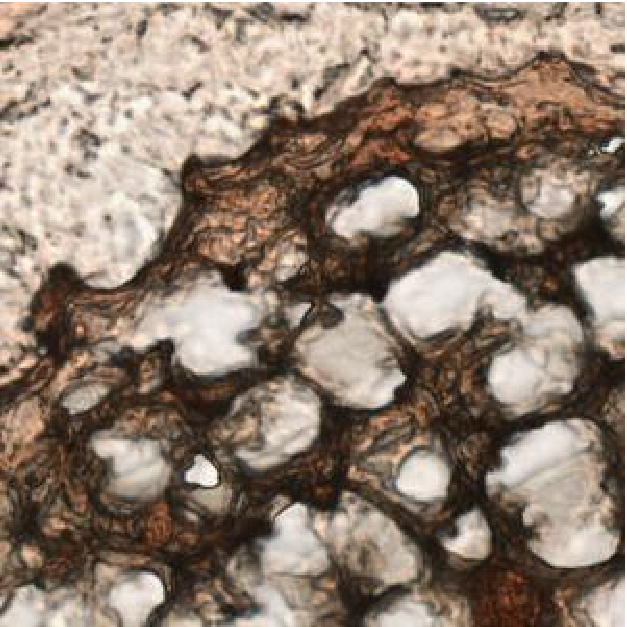}
            \caption{Original Image}
        \end{subfigure}
        \hspace{-0.02\textwidth}
        \begin{subfigure}{0.16\textwidth} 
            \centering
            \includegraphics[width=2.5cm, height=2.5cm]{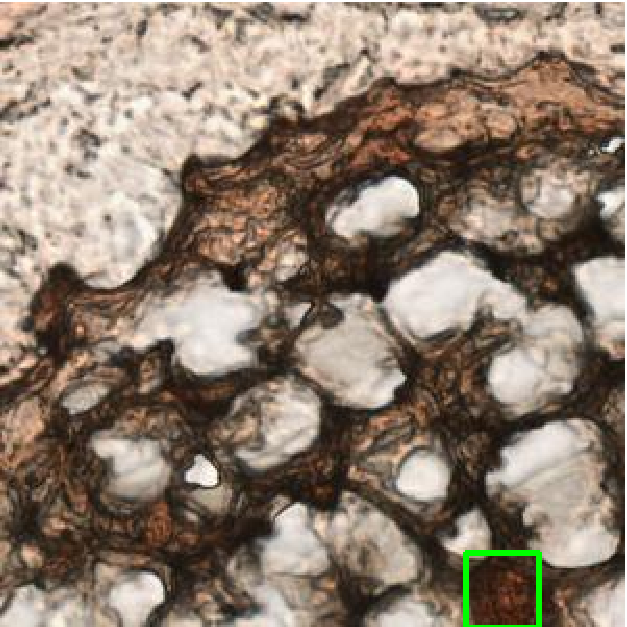}
            \caption{Ground Truth}
        \end{subfigure}
        \hspace{-0.02\textwidth}
        \begin{subfigure}{0.16\textwidth} 
            \centering
            \includegraphics[width=2.5cm, height=2.5cm]{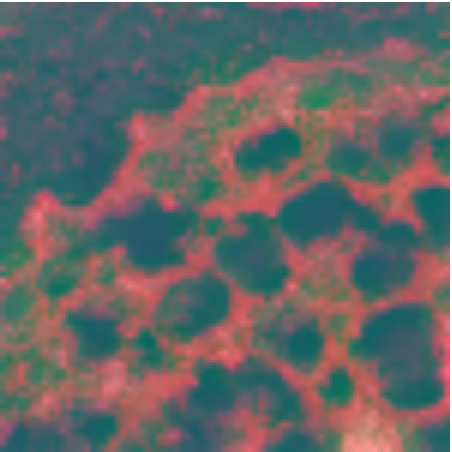}
            \caption{Layer 1}
        \end{subfigure}
        \hspace{-0.02\textwidth}
        \begin{subfigure}{0.16\textwidth} 
            \centering
            \includegraphics[width=2.5cm, height=2.5cm]{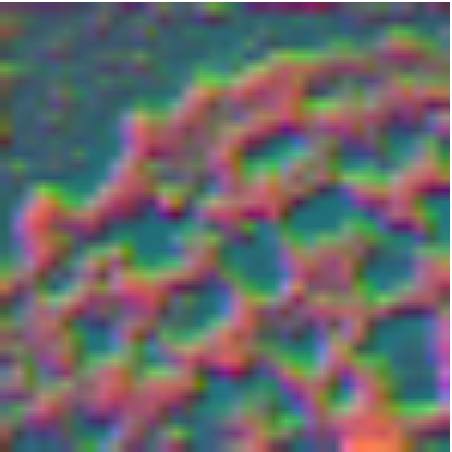}
            \caption{Layer 2}
        \end{subfigure}
        \hspace{-0.02\textwidth}
        \begin{subfigure}{0.16\textwidth} 
            \centering
            \includegraphics[width=2.5cm, height=2.5cm]{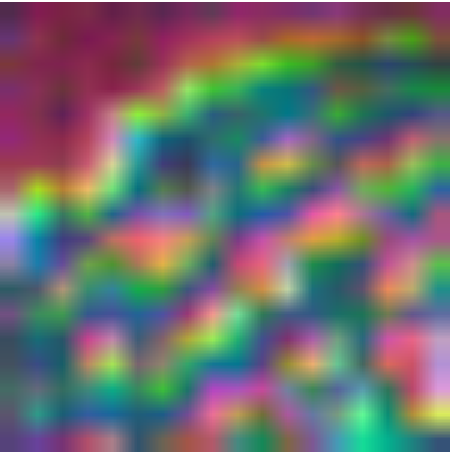}
            \caption{Layer 3}
        \end{subfigure}
        \hspace{-0.02\textwidth}
        \begin{subfigure}{0.16\textwidth} 
            \centering
            \includegraphics[width=2.5cm, height=2.5cm]{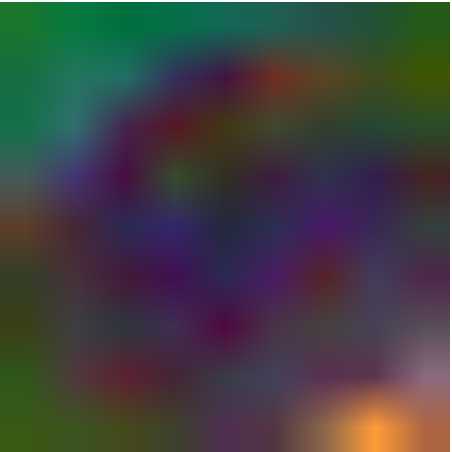}
            \caption{Layer 4}
        \end{subfigure} 
    \end{tabular}

    \captionsetup{justification=justified, singlelinecheck=false, format=hang}
    \caption{Feature analysis comparison between ViT (top row) and EI-ViT (bottom row) architectures, showing the original image, ground truth bounding box of the tumorous cell, and feature maps across 4 stages.}
    \label{fig:feature-analysis-pca}
\end{figure*}

\textbf{Feature Analysis:}. We conducted a detailed analysis of the feature maps generated by the baseline ViT architecture and its enhanced counterpart with our interaction modules on feature representations within the CCellBio dataset to evaluate the impact of our proposed enhancements. Using Principal Component Analysis (PCA), we analyzed the output feature maps from four transformer blocks of both the baseline ViT and EI-ViT model, shown in Figure \ref{fig:feature-analysis-pca}. These Transformer stages are designed to compute features at different spatial resolutions, providing multi-scale feature representations. Note that in each stage, there are multiple transformer blocks. In our setup, we decided to use the plain ViT architecture which contains 3 blocks per stage.

\begin{figure*}[htpb]
    \centering
    \begin{tabular}{c c}
        \multirow{2}{*}[1.2cm]{\rotatebox{90}{\textbf{ViT}}} &
        \begin{subfigure}{0.16\textwidth} 
            \centering
            \includegraphics[width=2.5cm, height=2.5cm]{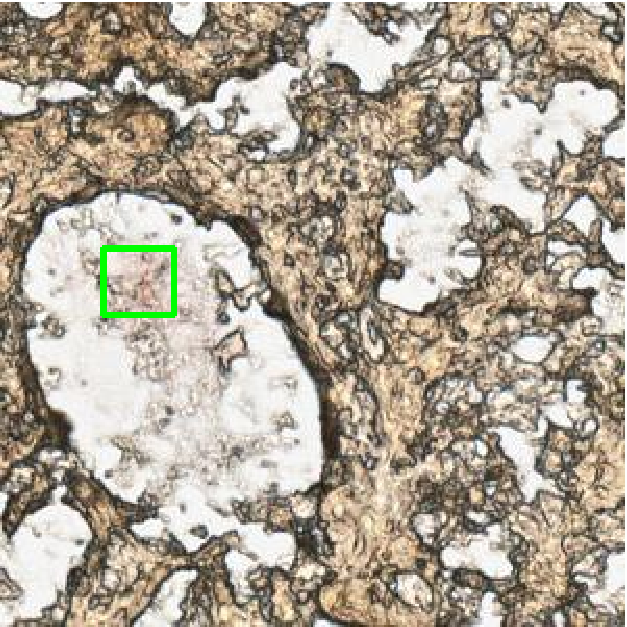}
        \end{subfigure}
        \hspace{-0.02\textwidth}
        \begin{subfigure}{0.16\textwidth} 
            \centering
            \includegraphics[width=2.5cm, height=2.5cm]{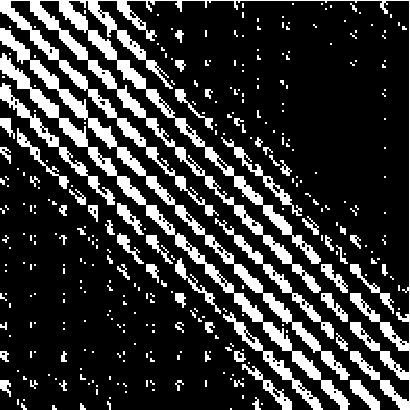}
        \end{subfigure}
        \hspace{-0.02\textwidth}
        \begin{subfigure}{0.16\textwidth} 
            \centering
            \includegraphics[width=2.5cm, height=2.5cm]{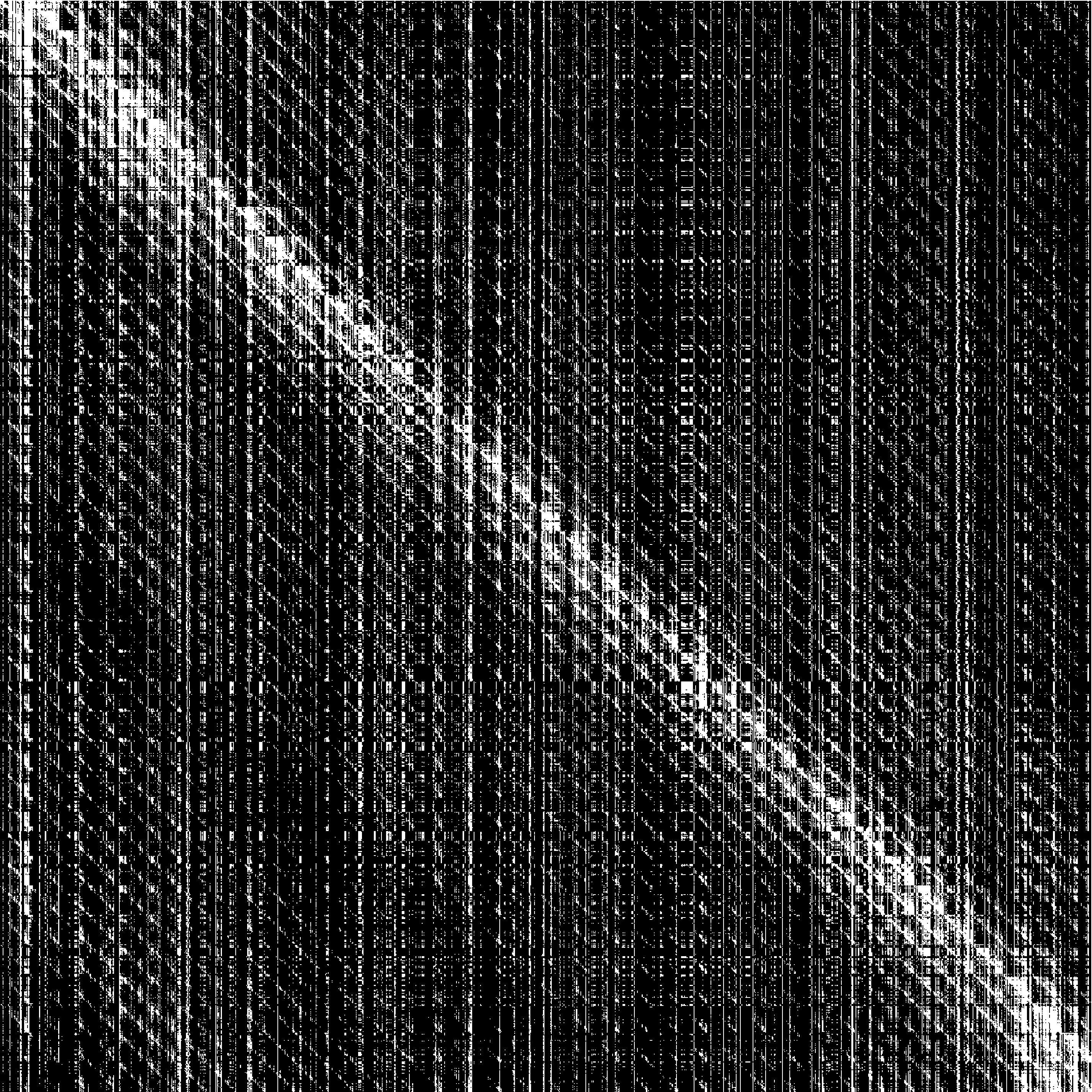}
        \end{subfigure}
        \hspace{-0.02\textwidth}
        \begin{subfigure}{0.16\textwidth} 
            \centering
            \includegraphics[width=2.5cm, height=2.5cm]{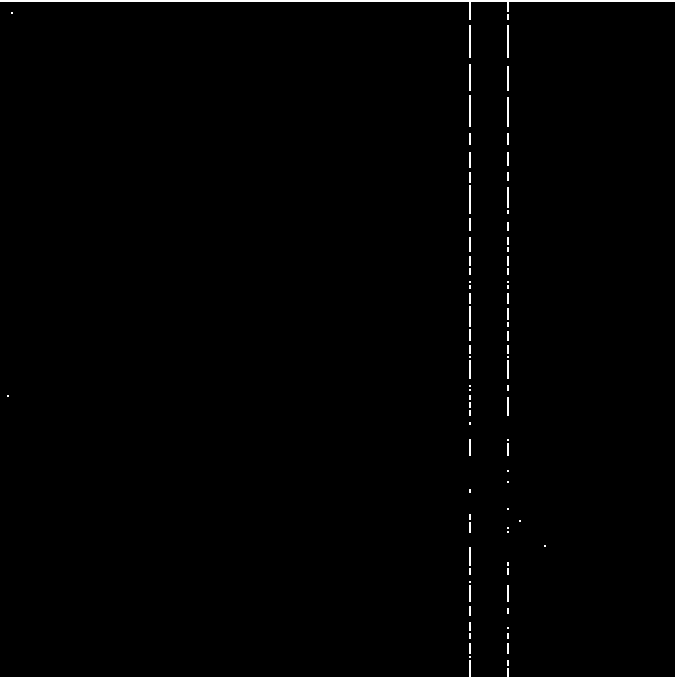}
        \end{subfigure}
        \hspace{-0.02\textwidth}
        \begin{subfigure}{0.16\textwidth} 
            \centering
            \includegraphics[width=2.5cm, height=2.5cm]{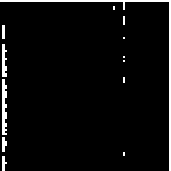}
        \end{subfigure}
        \hspace{-0.02\textwidth}
        \begin{subfigure}{0.16\textwidth} 
            \centering
            \includegraphics[width=2.5cm, height=2.5cm]{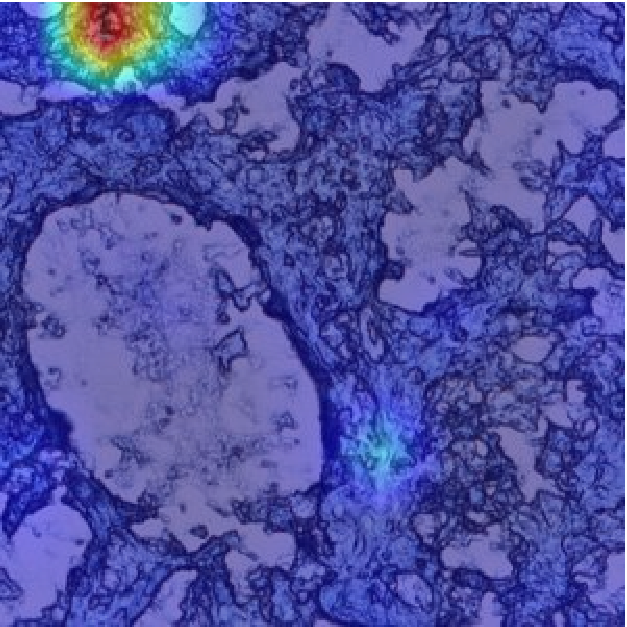}
        \end{subfigure} 
        \\
        \multirow{2}{*}[2.1cm]{\rotatebox{90}{\textbf{EI-ViT}}} &
        \begin{subfigure}{0.16\textwidth} 
            \centering
            \includegraphics[width=2.5cm, height=2.5cm]{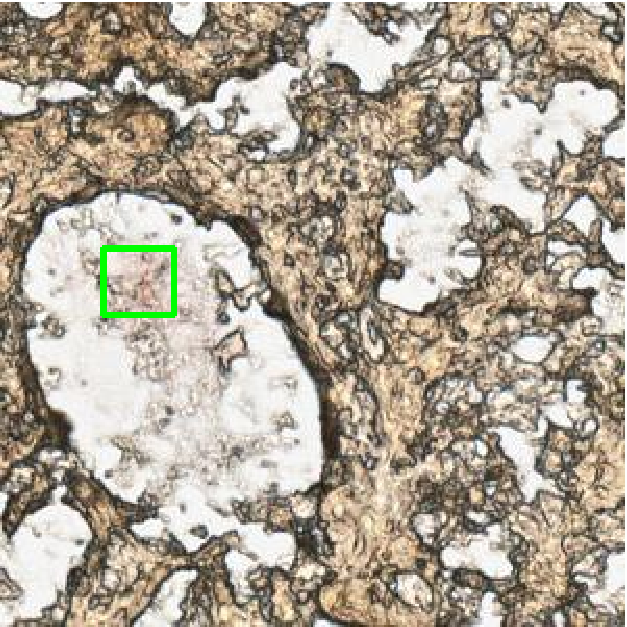}
            \caption{Ground Truth}
        \end{subfigure}
        \hspace{-0.02\textwidth}
        \begin{subfigure}{0.16\textwidth} 
            \centering
            \includegraphics[width=2.5cm, height=2.5cm]{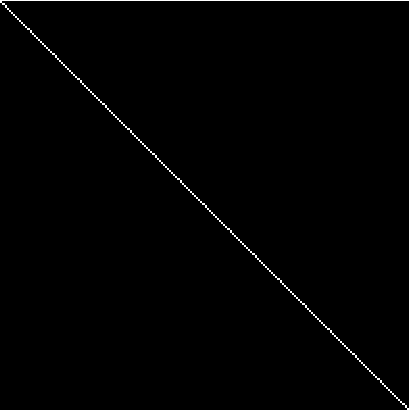}
            \caption{Attention B1}
        \end{subfigure}
        \hspace{-0.02\textwidth}
        \begin{subfigure}{0.16\textwidth} 
            \centering
            \includegraphics[width=2.5cm, height=2.5cm]{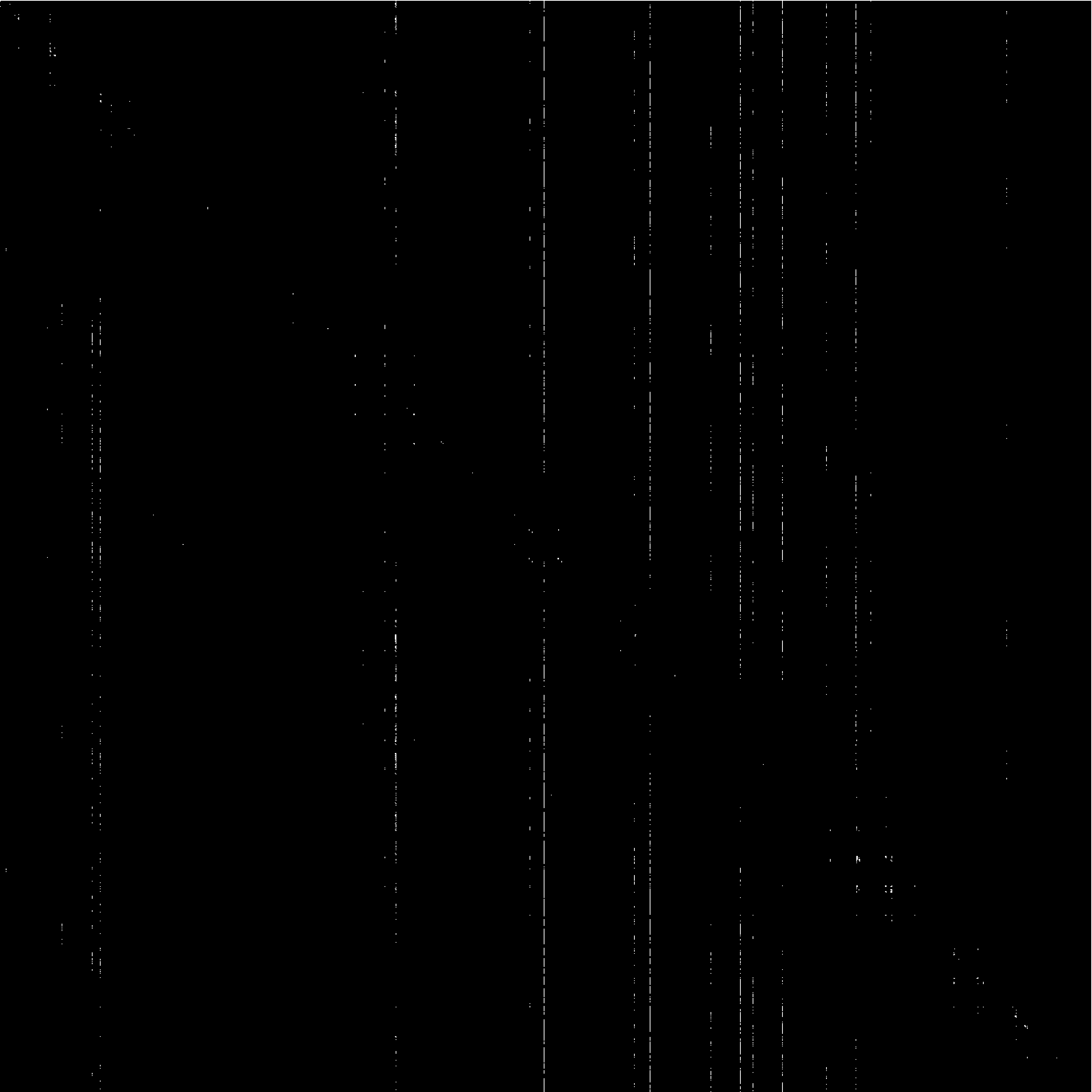}
            \caption{Attention B5}
        \end{subfigure}
        \hspace{-0.02\textwidth}
        \begin{subfigure}{0.16\textwidth} 
            \centering
            \includegraphics[width=2.5cm, height=2.5cm]{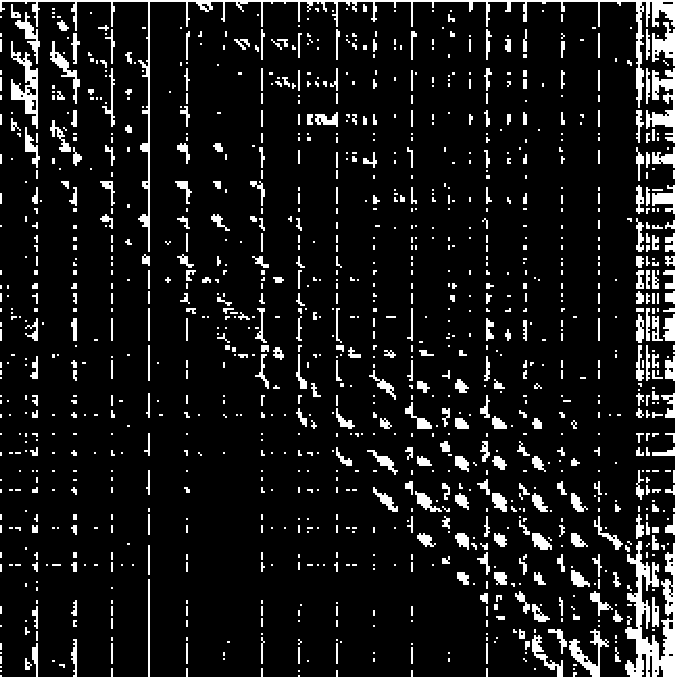}
            \caption{Attention B8}
        \end{subfigure}
        \hspace{-0.02\textwidth}
        \begin{subfigure}{0.16\textwidth} 
            \centering
            \includegraphics[width=2.5cm, height=2.5cm]{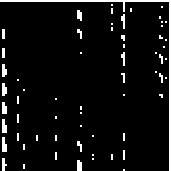}
            \caption{Attention B12}
        \end{subfigure}
        \hspace{-0.02\textwidth}
        \begin{subfigure}{0.16\textwidth} 
            \centering
            \includegraphics[width=2.5cm, height=2.5cm]{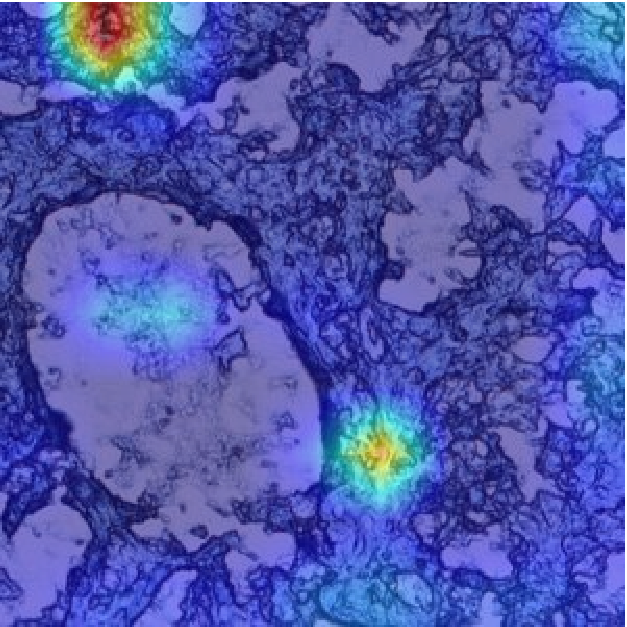}
            \caption{Proj Attention}
        \end{subfigure} 
    \end{tabular}

    \caption{Attention map analysis comparison between ViT (top row) and EI-ViT (bottom row) architectures. The attention maps shown here are enhanced  using an Otsu Thresholding algorithm for illustrative purposes. For both models, (a) shows the input image with the ground truth bounding box. (b), (c), (d), and (e) display the attention maps at blocks 1, 5, 8, and 12, respectively. (f) illustrates the projected final attention map.}
    \label{fig:attention-map-analysis}
\end{figure*}

\vspace{10pt}
\noindent The analysis reveals significant differences in the feature maps that align with the performance variations between the two models. Notably, the enhanced interaction prior to self-attention in the EI-ViT model produces sharper and more detailed feature maps with focused representations of objects. This is evident in the feature maps shown in subfigures (c), (d), (e), and (f) of Figure \ref{fig:feature-analysis-pca}, where the EI-ViT model displays more intricate textures compared to the baseline ViT.

\vspace{10pt}
\noindent This observation supports our hypothesis that enabling feature token interactions before self-attention adds complexity to the features, making them more distinct and representative of their true semantic classes. For instance, in the final feature map layer, as depicted in subfigure (f) of both models, the EI-ViT demonstrates a much more condensed and focused representation on the true object compared to the more diffuse focus of the baseline ViT. This behavior was consistently observed across multiple instances in our feature analysis, underscoring the effectiveness of the interaction modules in refining feature representations.

\vspace{10pt}
\noindent We perform Center Kernel Alignment (CKA) similarity analysis ~\cite{DBLP:journals/corr/abs-1905-00414} to quantify the feature similarity between ViT and EI-ViT. This analysis aims to investigate how the enhanced interaction modules affect the similarity of features at different layers. We evaluate the similarity of the feature maps produced by the ViT and EI-ViT backbones across four levels using the CCellBio test set. Note that each level contains multiple blocks, with a total of 12 blocks in the ViT model, subdivided into 4 levels, each consisting of 3 blocks. We examine the feature maps at these 4 levels.

\begin{figure}[t]
    \centering
    \begin{minipage}{0.45\textwidth}
        \centering
        \includegraphics[width=\linewidth]{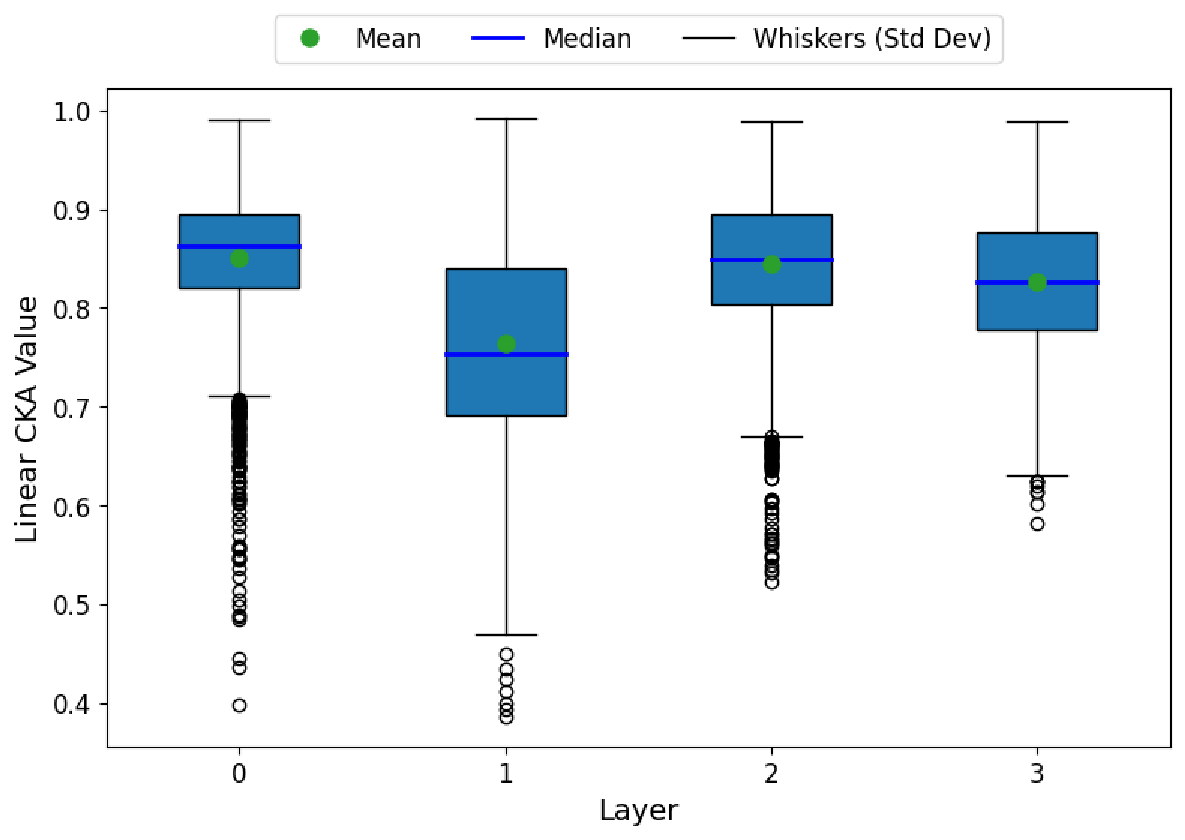}
        \caption{Linear CKA Similarity for the feature maps at different levels for both ViT and EI-ViT backbones.}
        \label{fig:linear-cka}
    \end{minipage}
    \hfill
    \begin{minipage}{0.45\textwidth}
        \centering
        \includegraphics[width=\linewidth]{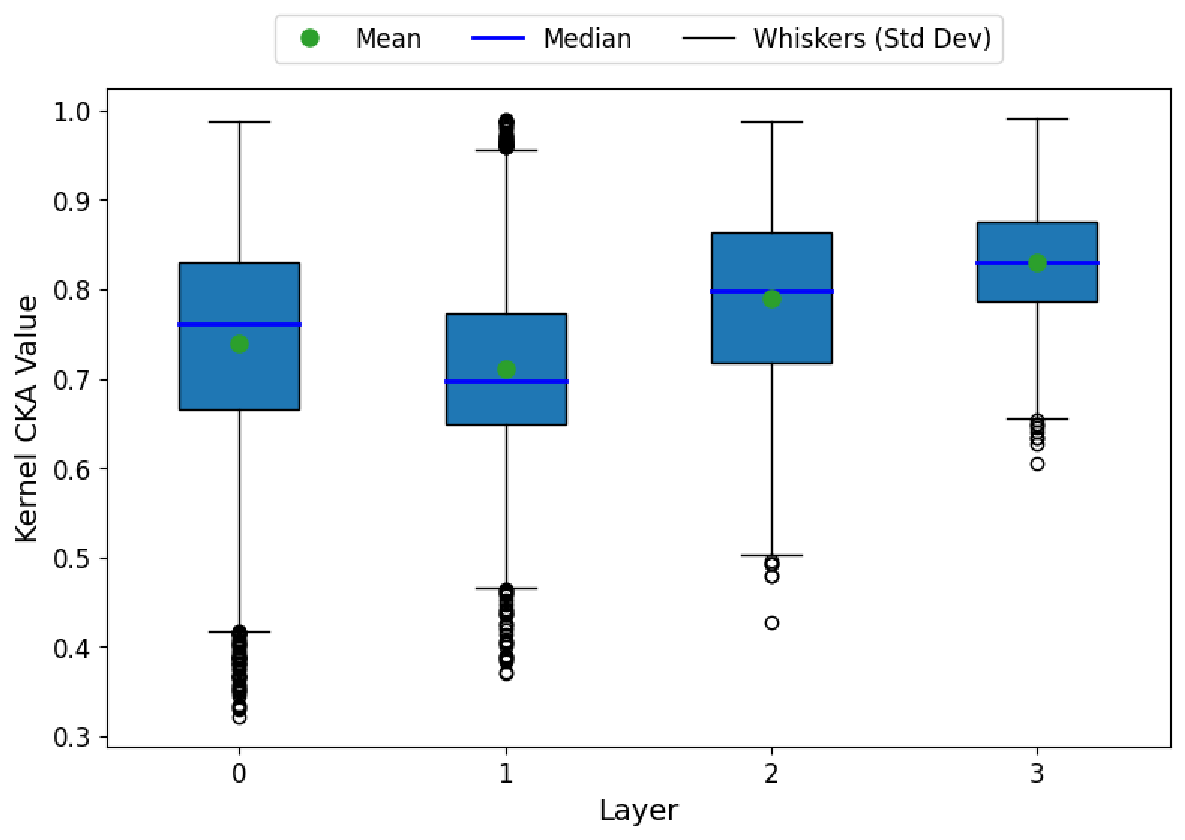}
        \caption{Kernel CKA Similarity for the feature maps at different levels for both ViT and EI-ViT backbones.}
        \label{fig:kernel-cka}
    \end{minipage}
\end{figure}

\vspace{10pt}
\noindent For each data point (test image), we compute both linear CKA and kernel CKA to measure feature similarity. The results of the CKA analysis, including the mean, median, and standard deviation, are shown in Figures~\ref{fig:linear-cka} and~\ref{fig:kernel-cka}. For the linear CKA analysis, the mean similarity values obtained for the four stages (1 to 4) were \( 0.8513 \pm 0.0650 \), \( 0.7650 \pm 0.1012 \), \( 0.0713 \pm 0.5220 \), and \( 0.8263 \pm 0.0705 \), respectively. In the case of kernel CKA, the corresponding similarity scores were \( 0.7401 \pm 0.1187 \), \( 0.7105 \pm 0.1017 \), \( 0.7898 \pm 0.0978 \), and \( 0.8306 \pm 0.0644 \), respectively. In both linear and kernel CKA comparisons, we observe that while the features exhibit a high degree of similarity, notable differences remain. Specifically, the CKA scores for the last layer are consistently lower than 0.86 for both linear and kernel CKA, indicating some divergence in the feature representations. Furthermore, there are significant deviations and outliers, with a tendency towards less similarity in some cases. Incorporating  ACP and CAT enables the EI-ViT backbone to learn features that differ from those learned by the baseline ViT. This divergence is reflected in both of the CKA and PCA analysis, where the enhanced model demonstrates a more diverse set of feature representations.
 
\vspace{10pt}
\noindent \emph{Attention Map Analysis:} We analyze the attention maps produced by ViT and EI-ViT to investigate the impact of the enhanced interaction modules on the multi-head attention mechanism. In Figure~\ref{fig:attention-map-analysis},  we extract and present the attention maps from blocks $1$, $5$, $8$, and $12$, applying a pattern-preserving Otsu Thresholding algorithm to enhance the visibility as attention map patterns are often difficult to see and interpret. Our analysis  highlights the influence of the interaction modules on the attention behavior of the vision transformer. In the early layers of the standard ViT architecture, the attention maps exhibit strong diagonal patterns, as shown in Figure~\ref{fig:attention-map-analysis} (a). This behavior is typical for architectures like ViT which tend to focus on localized patterns in early layers during early processing stages, and suggest that the network initially adopts locality and inductive biases similar to those seen in convolutional filters. In contrast, the EI-ViT model, which incorporates interaction modules, shows significantly less attention activity in the early layers, Figure~\ref{fig:attention-map-analysis} (a) and (b). This behavior indicates that the EI-ViT model does not prioritize learning inductive biases and locality in the initial processing stages, deviating from the more traditional ViT approach that builds on these biases.

As the network advances to deeper layers, a shift in behavior becomes evident. In the ViT model, the later layers, such as those shown in Figure~\ref{fig:attention-map-analysis} (d) and (e), exhibit reduced attention intensity compared to early layers. Thus the network's ability to capture global relationships diminishes in later layers. In contrast, the attention maps of EI-ViT in Figure~\ref{fig:attention-map-analysis} present a more diverse and focused distribution, particularly in the deeper layers. Figure~\ref{fig:attention-map-analysis}  (e) highlights more distinct vertical attention patterns for EI-ViT, as opposed to the more generalized focus seen in ViT. This shift suggests that the enhanced interaction modules in EI-ViT allow the network to distribute its attention across a broader set of regions in deeper feature layers, enabling it to focus on finer details within stronger semantic feature maps. As a result, EI-ViT is better equipped to capture and distinguish objects across various semantic classes, even when their appearances are similar.

\section{Ablation Study}

To understand the individual contributions of  aggressive convolutional pooling and  conceptual attention transformation in enhancing the
model's detection capabilities, we evaluate the model's performance using each component in isolation, We assess how the interactions facilitated by each influence overall performance. Additionally, we conduct experiments by varying the number of convolutional pooling blocks and the number of concepts independently. In this experiment, we focus on the ViT architecture, as it represents the foundational transformer model, and evaluate its performance on the CCellBio dataset. Using ViT as the baseline provides a clear framework for comparison, allowing us to assess the effects of the introduced components within a well-established context.

\begin{figure}[H]
    \centering
    \includegraphics[width=\columnwidth]{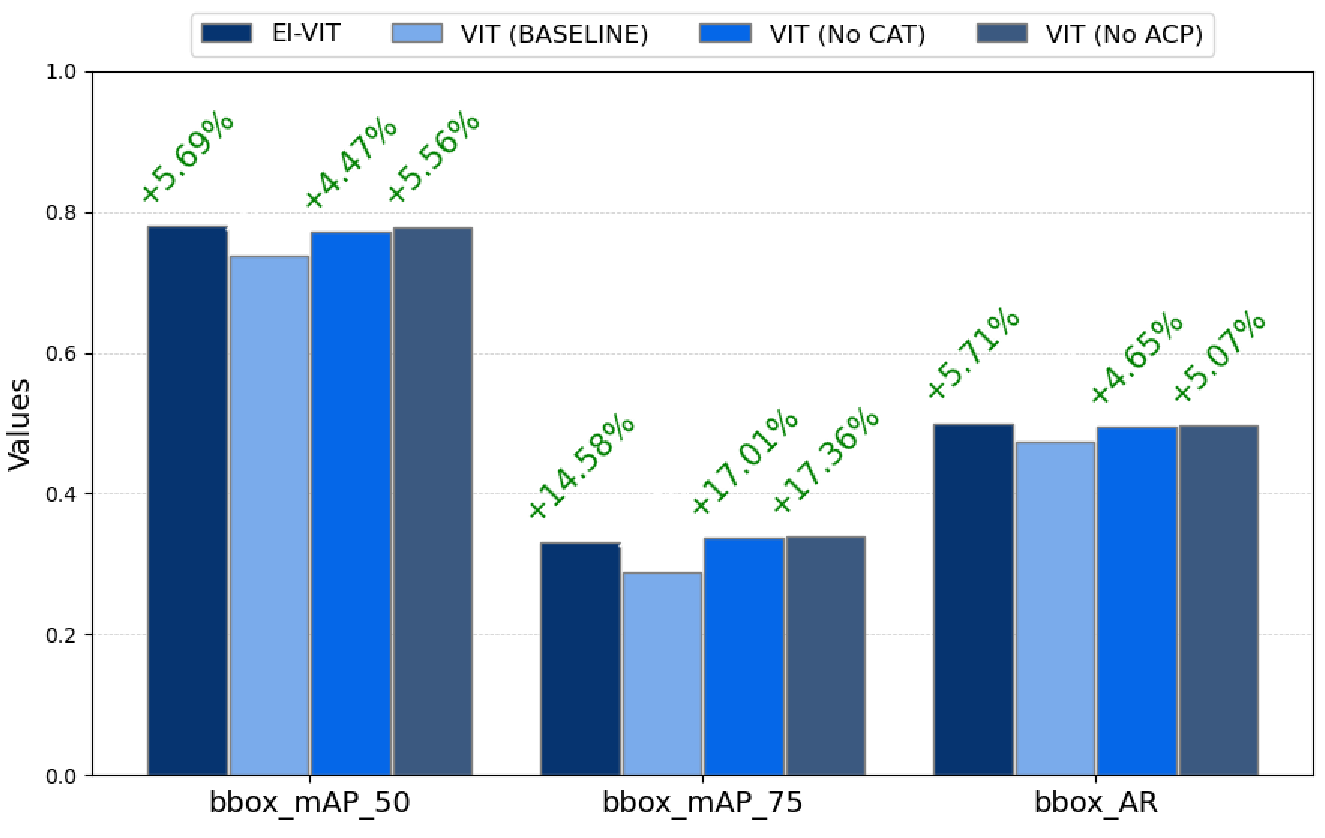}
    \caption{Isolation Benchmark: ViT (No CAT) and ViT (No ACP) represent the ViT architectures incorporating only the aggressive convolutional pooling (ACP) or the conceptual attention transformation (CAT), respectively. The plot illustrates the ablation benchmark for each component independently, showcasing how each component influences the performance of the model when tested in isolation.}
    \label{fig:ablation-isolation}
\end{figure}

\vspace{10pt}
\noindent  \textbf{Isolation Assessment.} We begin by training the ViT model with either the ACP  layer or the CAT  independently to assess the contribution of each component to the overall performance. The recorded benchmarks are presented in Figure \ref{fig:ablation-isolation}, which shows the performances for mAP50, mAP75, and AR for the baseline ViT, EI-ViT, and ViT models with isolated ACP and CAT components. 

\vspace{10pt}
\noindent Our observations reveal that removing the aggressive convolutional pooling results in a 5.56\% increase in mAP50 and a 5.07\% increase in AR compared to the baseline ViT. However, compared to the EI-ViT model, this leads to a reduction of 0.13\% in mAP50 and 0.65\% in AR. Removing the conceptual attention transformer has a more noticeable impact, reducing mAP50 by 1.22\% and AR by 1.06\% compared to the EI-ViT model. When only the ACP layer is included, the mAP75 metric improves by 17.36\%. Similarly, including only the CAT results in a 17.01\% improvement in mAP75 compared to the baseline, outperforming the EI-ViT architecture as well. This is because competing features for larger objects extracted by the convolution and attention mechanisms when both components are present make it more challenging for the network to perform well on the mAP75 metric. Additionally, we conducted extra epoch training with both the ViT-CAT and ViT-ACP models. While the performance of the isolated aggressive convolutional pooling ViT-ACP did not improve further, we observed that the isolated ViT-CAT continued to lower losses and gained extra points in the benchmark, even outperforming the EI-ViT for the CCellBio dataset across all metrics. Specifically, there was a 1.83\% improvement in mAP, 0.13\% in mAP50, 4.55\% in mAP75, and 0.20\% in AR when compared to the EI-ViT, which contains both the convolutional pooling and conceptual attention transformer layers. Our studies reveal that at lower epoch training, the utilization of the ACP component helps the network converge faster and achieve high KPIs. However, when training for more epochs, the CAT component can learn more relevant features, removing the need for the ACP component and achieving the higher KPIs.

\noindent 

\begin{figure*}[htpb]
    \centering
    \begin{minipage}{0.45\textwidth}
        \centering
      \includegraphics[width=\columnwidth]{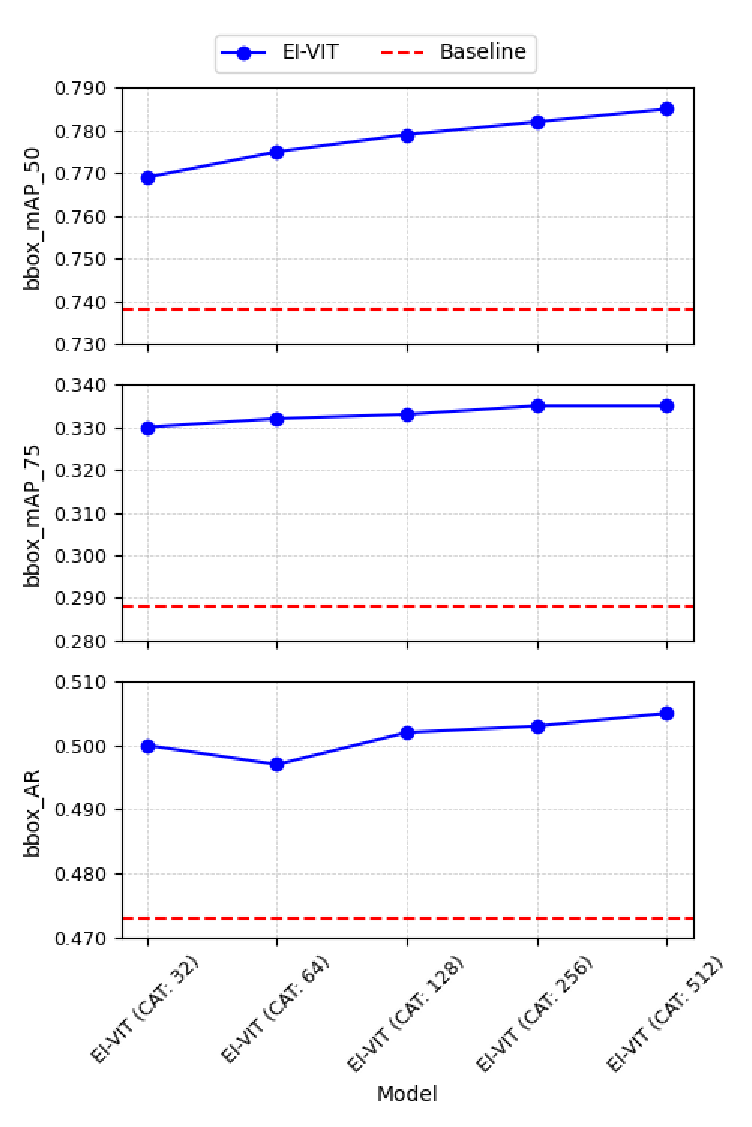}
    \caption{There is a consistent positive correlation between the number of concepts and performance across all three metrics (mAP50, mAP75, and AR).}
    \label{fig:ablation-cat}
    \end{minipage}\hfill
    \begin{minipage}{0.45\textwidth}
        \centering
         \includegraphics[width=\columnwidth]{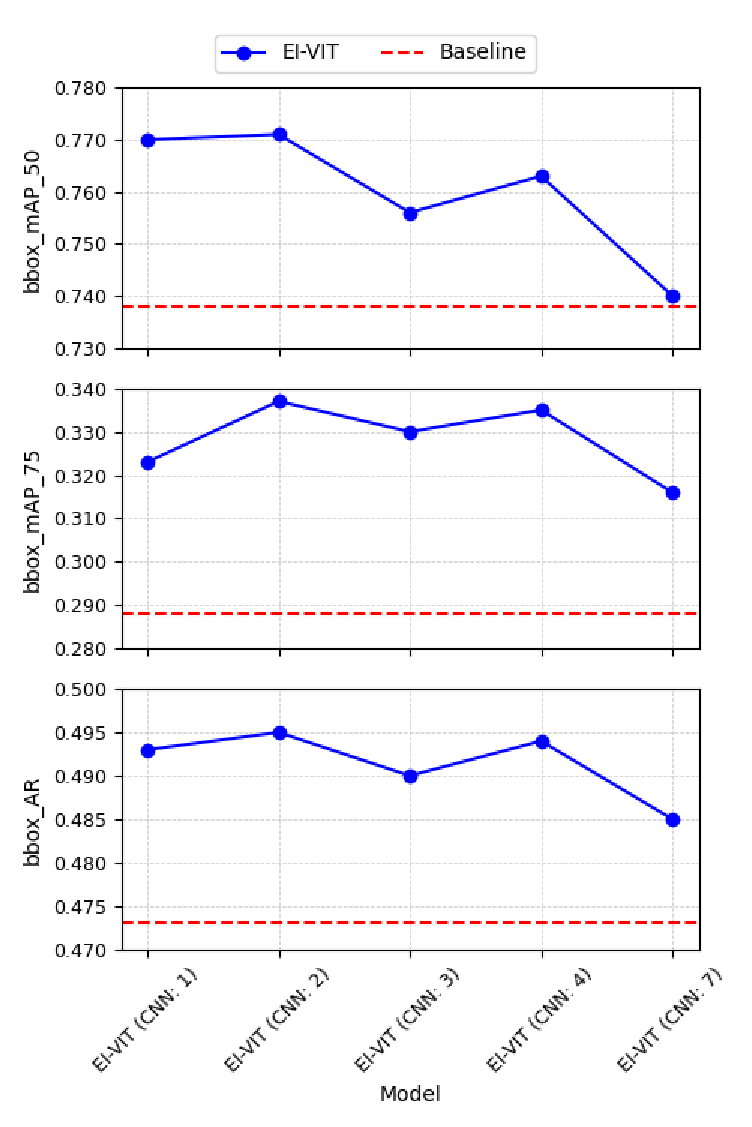}
    \caption{ When aggressive convolutional pooling is applied, the performance improves over the baseline across mAP50, mAP75, and AR.}
    \label{fig:ablation-cnn}
    \end{minipage}
\end{figure*}

\vspace{10pt}
\noindent \textbf{Aggressive Convolutional Pooling}. We explore the impact of varying the number of convolutional pooling layers on overall network performance. This technique is incorporated into the baseline ViT architecture, and the network's performance is evaluated using AR and mAP at two different thresholds, mAP50 and mAP75. The results shown in Figure \ref{fig:ablation-cnn} reveal a clear relationship between the number of convolutional pooling layers and the network's performance. When aggressive convolutional pooling is applied, there is a noticeable improvement in the mAP50 and AR, particularly when the number of pooling layers is kept relatively low. When the number of pooling layers is limited to four, the network improves its ability to locate objects and recall true positives. An intriguing observation shows that the mAP75 metric performance drops below the baseline when the number of layers is set to  $1$ or $7$, suggesting that too few or too many pooling layers hinder the model's ability to handle object localization at higher thresholds.

\vspace{10pt}
\noindent The best performance across all three metrics is achieved when the number of pooling layers is set to two. With this configuration, the network shows a relative improvement of 4.47\%, 17.01\%, and 4.23\% in mAP50, mAP75, and AR, respectively. This indicates that the optimal number of convolutional pooling layers lies in a balanced configuration, where pooling is applied strategically to extract essential features without overly distorting the network's capacity to retain fine-grain details.

\vspace{10pt}
\noindent  Despite the overall improvements seen with the introduction of aggressive convolutional pooling, performance peaks at two layers. Adding additional layers beyond this point results in diminishing returns. In fact, when more than two pooling layers are applied, the network's performance begins to decline. We hypothesize that this decline in performance is due to the aggressive application of pooling operations after every convolutional layer. This rapid pooling may cause the network to discard or overly simplify important spatial information, making it harder for the model to preserve the contextual details necessary for effective object detection and localization. Even with the use of residual connections, the frequent pooling operations may hinder the network's ability to learn discriminative features and retain relevant information across deeper layers. While aggressive convolutional pooling can improve the performance of the ViT architecture in certain cases, it is crucial to find the optimal balance between pooling layers. These findings underscore the importance of carefully controlling the amount of pooling applied, as it can have a substantial impact on the network's ability to generalize and perform well on object detection tasks.

\vspace{10pt}
\noindent\textbf{Conceptual Attention Transformation:} To evaluate the influence of the number of concepts on performance, we conduct a comprehensive analysis using the baseline ViT architecture, enhanced solely by the CAT layer. Performance metrics include mAP50, mAP75, and AR, with the number of concepts systematically varied from $32$ to $512$. The results are presented in Figure~\ref{fig:ablation-cat}. All three metrics (mAP50, mAP75, and AR) exhibit consistent and positive correlations between the number of concepts performance. As the number of concepts increase, the network shows a marked improvement in its ability to detect objects, with particular gains in recognizing smaller or more localized objects that require finer-grained attention. For mAP50, the performance boost is the most notable, with a relative improvement of up to 6.37\% when the number of concepts is increased to $512$. This suggests that a larger pool of concepts enables the model to capture more detailed objects features within the scene, leading to better localization and overall improved detection accuracy.

\vspace{10pt}
\noindent A similar pattern is observed for the mAP75 and AR metrics. At $512$ concepts, the network demonstrates relative performance gains of 16.32\% and 4.12\%, respectively, indicating that a comprehensive conceptual representation, provided by a higher number of concepts, yields more accurate results at higher intersection over Union (IoU) thresholds and improves recall.  These performance gains reflect  the model's ability to maintain high precision while simultaneously improving its recall, especially in cases where the object boundaries are more challenging to detect, or where a more comprehensive understanding of the object context is required.

\section{Limitations}\label{sec:limit}
 
\noindent In this section, we discuss limitations of our approach. We observed that the incorporation of the enhanced interaction mechanism may hinder the EI-DAT model's ability  to effectively learn deformable points. In this case, the query matrix is designed to learn and generate deformable points. These offset points are then used to interpolate feature maps. However, additional complexities added by the model enhancement make it difficult to compute queries accurately. This causes a decline in model performance that ultimately impacts its mAP performance as it struggles to  focus on relevant regions. 

\vspace{10pt}
\noindent We benchmark the performance of the enhanced interaction models without pretraining to demonstrate that the method does not rely on agnostic pretraining to achieve competitive results. Although the enhanced interaction models still outperform their respective baselines, we observe the models exhibit low mAP and AR scores when trained from random initialization on small-sized datasets. This is observed when the model has challenges achieving higher scores without pretrained weights when applied to the NIH Chest X-Ray dataset which has a limited size. 

\vspace{10pt}
\noindent Our ablation tests show that the current design does not allow for many CNN pooling layers. There are opportunities to explore ways to aggregate features  across different CNN pooling layers in a manor that creates larger receptive fields by increasing the number of effective CNN pooling layers.

\section{Conclusion} 
\noindent In this work, we introduced an enhanced interaction modeling approach for the vision-transformer backbone in object detectors. Our findings demonstrate that enabling interaction prior to self-attention improves performance across multiple challenging medical and concealed object detection datasets and for a diverse set of evaluation metrics. We showed that incorporating our aggressive attention pooling and conceptual attention transformation alters self-attention behavior by allowing it to learn more distinct features and attention maps when compared to original baseline models. The modified models distinguish and represent these features and attention maps more clearly in the feature space, even when objects of different classes have similar visual appearances. Such interactions make it easier for self-attention operations to focus on relevant and varied regions. Our experiments reveal that, with extended training epochs, the conceptual attention transformer does note require the ACP component to achieve competitive performance. In summary, this study demonstrates that prior self-attention interactions, both at local and global scales, are important complementary operations that enable the vision transformer to learn robust and diverse features for improved object detection.

\section{Acknowledgements}  We thank David Knowles and Peter Belhumeur for their constructive feedback. We also thank Afsana Salahudeen and John Drotos for help with data collection and literature reviews. This work was conducted at the Graphics Imaging and Light Measurement Lab (GILMLab) at Barnard College, Columbia University.

\bibliographystyle{ACM-Reference-Format}
\bibliography{ms.bib}

\begin{appendices}
\clearpage


\begin{table}[ht]
\centering
\caption{Parameter Configuration: ViT Backbone (ViT)}
\label{table:backbone_vit_configuration}
\begin{tabular}{ll}
\toprule
\textbf{Attribute}                   & \textbf{Value}                       \\ 
\midrule
Type                                 & ViT                               \\ 
Image Size                           & 512                                  \\ 
Patch Size                           & 4                                    \\ 
Embedding Dimension                  & 144, 288, 576, 1152                     \\ 
Down Factors                         & 2, 2, 2, 1                           \\ 
Number of Stages                     & 2, 3, 3, 3                           \\ 
Number of Attention Heads            & 8                                    \\ 
Drop Path Rate                       & 0.1                                  \\ 
Window Size                          & 14                                   \\ 
MLP Ratio                            & 4                                    \\ 
QKV Bias                             & True                                 \\ 
Normalization Configuration          & Layer Norm                           \\ 
Use Relative Positioning             & True                                 \\ 
\bottomrule
\end{tabular}
\end{table}

\begin{table}[ht]
\centering
\caption{Parameter Configuration: Enhanced Interaction for ViT Backbone (EI-ViT)}
\label{table:backbone_EI-VIT_configuration}
\begin{tabular}{ll}
\toprule
\textbf{Attribute}                   & \textbf{Value}                       \\ 
\midrule
Type                                 & EI-ViT                               \\ 
Image Size                           & 512                                  \\ 
Patch Size                           & 4                                    \\ 
Embedding Dimension                  & 48, 96, 192, 384                     \\ 
Down Factors                         & 2, 2, 2, 1                           \\ 
Global CNN Hidden Dimension          & 512                                  \\ 
Number of Stages                     & 2, 3, 3, 3                           \\ 
Number of Attention Heads            & 8                                    \\ 
Number of Concepts                   & 48, 96, 192, 384                     \\ 
Drop Path Rate                       & 0.1                                  \\ 
Window Size                          & 14                                   \\ 
MLP Ratio                            & 4                                    \\ 
QKV Bias                             & True                                 \\ 
Normalization Configuration          & Layer Norm                           \\ 
Use Relative Positioning             & True                                 \\ 
\bottomrule
\end{tabular}
\end{table}

\begin{table}[ht]
\centering
\caption{Parameter Configuration: Swin Backbone (Swin)}
\label{table:backbone_eccustomswin_configuration}
\begin{tabular}{ll}
\toprule
\textbf{Attribute}                   & \textbf{Value}                     \\ 
\midrule
Type                                 & Swin                               \\ 
Pretraining Image Size               & 512                                \\ 
Patch Size                           & 4                                  \\ 
Input Channels                       & 3                                  \\ 
Embedding Dimension                  & 288, 576, 1152, 2304                               \\ 
Depths                               & 2, 2, 6, 2                         \\ 
Number of Attention Heads            & 3, 6, 12, 24                       \\ 
Window Size                          & 7                                  \\ 
MLP Ratio                            & 4.0                                \\ 
QKV Bias                             & True                               \\ 
QK Scale                             & None                               \\ 
Drop Rate                            & 0.0                                \\ 
Attention Drop Rate                  & 0.0                                \\ 
Drop Path Rate                       & 0.2                                \\ 
Absolute Position Embedding (APE)    & False                              \\ 
Patch Normalization                  & True                               \\ 
Output Indices                       & 0, 1, 2, 3                         \\ 
Frozen Stages                        & -1                                 \\ 
\bottomrule
\end{tabular}
\end{table}

\begin{table}[ht]
\centering
\caption{Parameter Configuration: Enhanced Interaction for Swin Backbone (EI-Swin)}
\label{table:backbone_eccustomswin_configuration}
\begin{tabular}{ll}
\toprule
\textbf{Attribute}                   & \textbf{Value}                     \\ 
\midrule
Type                                 & EI-Swin                            \\ 
Pretraining Image Size               & 512                                \\ 
Patch Size                           & 4                                  \\ 
Input Channels                       & 3                                  \\ 
Embedding Dimension                  & 96, 192, 384, 768                                \\ 
Depths                               & 2, 2, 6, 2                         \\ 
Number of Attention Heads            & 3, 6, 12, 24                       \\ 
Window Size                          & 7                                  \\ 
MLP Ratio                            & 4.0                                \\ 
QKV Bias                             & True                               \\ 
QK Scale                             & None                               \\ 
Drop Rate                            & 0.0                                \\ 
Attention Drop Rate                  & 0.0                                \\ 
Drop Path Rate                       & 0.2                                \\ 
Absolute Position Embedding (APE)    & False                              \\ 
Patch Normalization                  & True                               \\ 
Output Indices                       & 0, 1, 2, 3                         \\ 
Frozen Stages                        & -1                                 \\ 
\bottomrule
\end{tabular}
\end{table}

\begin{table}[ht]
\centering
\caption{Parameter Configuration: DAT++ Backbone (DAT++)}
\label{table:backbone_configuration}
\begin{tabular}{ll}
\toprule
\textbf{Attribute}                   & \textbf{Value}               \\ 
\midrule
Type                                 & EI-DAT                       \\ 
Image Size                           & 512                          \\ 
Patch Size                           & 4                            \\ 
Expansion                            & 4                            \\ 
Dimension Stem                       & 64                           \\ 
\midrule
\textbf{Architecture}                &                              \\ 
\quad Dimensions                     & 128, 256, 512, 1024            \\ 
\quad Depths                         & 2, 4, 18, 2                  \\ 
\quad Heads                          & 2, 4, 8, 16                  \\ 
\quad Query Heads                    & 6, 12, 24, 48                \\ 
\quad Window Sizes                   & 7, 7, 7, 7                   \\ 
\midrule
\textbf{Drop Rates}                  &                              \\ 
\quad Attention Drop Rate            & 0.0                          \\ 
\quad Drop Path Rate                 & 0.0                          \\ 
\midrule
\textbf{Strides and Offsets}         &                              \\ 
\quad Strides                        & 8, 4, 2, 1                   \\ 
\quad Offset Range Factor            & -1, -1, -1, -1               \\ 
\quad Local Offset Range Factor      & -1, -1, -1, -1               \\ 
\quad Local Key-Value Sizes          & -1, -1, -1, -1               \\ 
\quad Offset Positional Embeddings   & F, F, F, F                   \\ 
\midrule
\textbf{Stage Specification}         &                              \\ 
\quad Stage 1                        & N, D                         \\ 
\quad Stage 2                        & N, D, N, D                   \\ 
\quad Stage 3                        & N, D (x18)                   \\ 
\quad Stage 4                        & D, D                         \\ 
\midrule
\textbf{Positional Embeddings and Groups} &                         \\ 
\quad Groups                         & 1, 2, 4, 8                   \\ 
\quad Use Positional Embeddings      & T, T, T, T                   \\ 
\quad Depthwise Conv Pos Embeddings  & F, F, F, F                   \\ 
\quad Scaling Ratios                 & 8, 4, 2, 1                   \\ 
\midrule
\textbf{Advanced Features}           &                              \\ 
\quad Dense-Wise MLPs                & T, T, T, T                   \\ 
\quad Kernel Sizes                   & 9, 7, 5, 3                   \\ 
\quad Query-Neighbor Kernel Sizes    & 3, 3, 3, 3                   \\ 
\quad Number of Queries              & 2, 2, 2, 2                   \\ 
\quad Query-Neighbor Activation      & exp                          \\ 
\quad Deform Groups                  & 0, 0, 0, 0                   \\ 
\quad NAT Kernel Sizes               & 7, 7, 7, 7                   \\ 
\quad Layer Scale Values             & -1, -1, -1, -1               \\ 
\quad Use LPUs                       & T, T, T, T                   \\ 
\midrule
\textbf{Output and Initialization}   &                              \\ 
\quad Output Indices                 & 1, 2, 3                      \\ 
\quad Number of Concepts             & 64, 128, 256, 512            \\ 
\bottomrule
\end{tabular}
\end{table}

\begin{table}[ht]
\centering
\caption{Parameter Configuration: Enhanced Interaction DAT++ Backbone (EI-DAT)}
\label{table:backbone_configuration}
\begin{tabular}{ll}
\toprule
\textbf{Attribute}                   & \textbf{Value}               \\ 
\midrule
Type                                 & EI-DAT                       \\ 
Image Size                           & 512                          \\ 
Patch Size                           & 4                            \\ 
Expansion                            & 4                            \\ 
Dimension Stem                       & 64                           \\ 
\midrule
\textbf{Architecture}                &                              \\ 
\quad Dimensions                     & 64, 128, 256, 512            \\ 
\quad Depths                         & 2, 4, 18, 2                  \\ 
\quad Heads                          & 2, 4, 8, 16                  \\ 
\quad Query Heads                    & 6, 12, 24, 48                \\ 
\quad Window Sizes                   & 7, 7, 7, 7                   \\ 
\midrule
\textbf{Drop Rates}                  &                              \\ 
\quad Attention Drop Rate            & 0.0                          \\ 
\quad Drop Path Rate                 & 0.0                          \\ 
\midrule
\textbf{Strides and Offsets}         &                              \\ 
\quad Strides                        & 8, 4, 2, 1                   \\ 
\quad Offset Range Factor            & -1, -1, -1, -1               \\ 
\quad Local Offset Range Factor      & -1, -1, -1, -1               \\ 
\quad Local Key-Value Sizes          & -1, -1, -1, -1               \\ 
\quad Offset Positional Embeddings   & F, F, F, F                   \\ 
\midrule
\textbf{Stage Specification}         &                              \\ 
\quad Stage 1                        & N, D                         \\ 
\quad Stage 2                        & N, D, N, D                   \\ 
\quad Stage 3                        & N, D (x18)                   \\ 
\quad Stage 4                        & D, D                         \\ 
\midrule
\textbf{Positional Embeddings and Groups} &                         \\ 
\quad Groups                         & 1, 2, 4, 8                   \\ 
\quad Use Positional Embeddings      & T, T, T, T                   \\ 
\quad Depthwise Conv Pos Embeddings  & F, F, F, F                   \\ 
\quad Scaling Ratios                 & 8, 4, 2, 1                   \\ 
\midrule
\textbf{Advanced Features}           &                              \\ 
\quad Dense-Wise MLPs                & T, T, T, T                   \\ 
\quad Kernel Sizes                   & 9, 7, 5, 3                   \\ 
\quad Query-Neighbor Kernel Sizes    & 3, 3, 3, 3                   \\ 
\quad Number of Queries              & 2, 2, 2, 2                   \\ 
\quad Query-Neighbor Activation      & exp                          \\ 
\quad Deform Groups                  & 0, 0, 0, 0                   \\ 
\quad NAT Kernel Sizes               & 7, 7, 7, 7                   \\ 
\quad Layer Scale Values             & -1, -1, -1, -1               \\ 
\quad Use LPUs                       & T, T, T, T                   \\ 
\midrule
\textbf{Output and Initialization}   &                              \\ 
\quad Output Indices                 & 1, 2, 3                      \\ 
\bottomrule
\end{tabular}
\end{table}

\end{appendices}

\end{document}